%% file: main.tex
\documentclass[10pt,twocolumn,letterpaper]{article}
\usepackage[pagenumbers]{cvpr}

\input{preamble}
\definecolor{cvprblue}{rgb}{0.21,0.49,0.74}
\usepackage[pagebackref,breaklinks,colorlinks,allcolors=cvprblue]{hyperref}
\def\confName{}
\def\confYear{}

\title{Lost in Read-out: Locating the Image Quality Gap\\in Vision-Language Models with a Closed-Form Edit}

\author{
Zhuangzi Li$^{1}$ \quad
Shilv Cai $^{1}$ \thanks{Corresponding author.} \quad
Jian Jin $^{1}$ \quad
Zhiyi Pan $^{2}$ \quad
Haihong Xiao $^{3}$ \quad
Weisi Lin $^{1}$
\\
$^{1}$ Nanyang Technological University, Singapore\\
$^{2}$Tianjin University, Tianjin, China\\
$^{3}$Hefei University of Technology, Hefei, China\\
}

\begin{document}
\maketitle
\input{fig/teaser}
\input{sec/0_abstract}
\input{sec/1_intro}
\input{sec/2_related}

\input{sec/3_gap}
\input{sec/5_method}
\input{sec/6_experiments}
\input{sec/7_costs}
\input{sec/8_conclusion}
\input{sec/9_llm_statement}
{
    \small
    \bibliographystyle{ieeenat_fullname}
    \bibliography{main}
}

\clearpage
\setcounter{section}{0}
\setcounter{figure}{0}
\setcounter{table}{0}
\setcounter{equation}{0}
\renewcommand{\thesection}{S\arabic{section}}
\renewcommand{\thetable}{S\arabic{table}}
\renewcommand{\thefigure}{S\arabic{figure}}
\renewcommand{\theequation}{S\arabic{equation}}
\input{sec/X_suppl}

\end{document}

%% file: preamble.tex
\usepackage{cuted} %

\usepackage{currfile} %

\usepackage{caption} %

\makeatletter
\renewcommand{\paragraph}{\@startsection{paragraph}{4}{\z@}{0.8ex \@plus 0.3ex \@minus 0.2ex}{-1em}{\normalfont\normalsize\bfseries}}
\makeatother

\usepackage{multirow}
\usepackage{tikz}
\usetikzlibrary{positioning,arrows.meta,fit,calc,backgrounds,decorations.pathreplacing,shapes.misc,shapes.geometric}

%% file: fig/teaser.tex
\definecolor{visc}{RGB}{66,133,200}
\definecolor{lmc}{RGB}{150,150,150}
\definecolor{hotc}{RGB}{230,120,30}
\definecolor{editc}{RGB}{40,150,90}
\definecolor{panelc}{RGB}{246,246,246}
\begin{strip}
  \centering
  \begin{tikzpicture}[
      font=\footnotesize,
      blk/.style={draw=#1!80!black, fill=#1!25, minimum width=0.17cm, minimum height=0.62cm, inner sep=0pt},
      panel/.style={draw=black!20, fill=panelc, rounded corners=3pt},
      ttl/.style={font=\small\bfseries, anchor=north west},
      arr/.style={-{Stealth[length=4pt]}, thick},
      agent/.style={circle, draw=black!60, fill=white, inner sep=0pt, minimum size=0.42cm, font=\scriptsize\sffamily},
      take/.style={draw=black!15, fill=white, rounded corners=2pt, minimum width=5.35cm, minimum height=0.72cm, align=center, font=\scriptsize, inner sep=2pt},
      lab/.style={font=\scriptsize, inner sep=1pt},
    ]
    \fill[panel] (0,0) rectangle (5.65,4.3);
    \fill[panel] (5.85,0) rectangle (11.5,4.3);
    \fill[panel] (11.7,0) rectangle (17.35,4.3);
    \node[ttl] at (0.1,4.25) {(a) Decodable, but not said};
    \node[ttl] at (5.95,4.25) {(b) Where it is lost};
    \node[ttl] at (11.8,4.25) {(c) Rewrite the read-out};

    \shade[left color=yellow!60!orange, right color=visc!60, draw=black!50] (0.2,2.65) rectangle (0.8,3.25);
    \node[lab, anchor=north] at (0.5,2.56) {image};
    \foreach \i in {0,...,6} {
      \pgfmathsetmacro{\x}{1.05+\i*0.25}
      \ifnum\i=2
        \node[blk=hotc] (v\i) at (\x,2.95) {};
      \else
        \node[blk=visc] (v\i) at (\x,2.95) {};
      \fi
    }
    \node[lab, anchor=north, text=visc!80!black] at (1.8,2.56) {vision encoder};
    \foreach \i in {0,...,5} {
      \pgfmathsetmacro{\x}{2.95+\i*0.25}
      \node[blk=lmc] (l\i) at (\x,2.95) {};
    }
    \node[lab, anchor=north, text=black!60] at (3.58,2.56) {language model};
    \draw[arr] (0.82,2.95) -- (v0.west);
    \draw[arr] (v6.east) -- (l0.west);
    \node[draw=black!50, fill=white, rounded corners=2pt, align=center, font=\scriptsize, inner sep=2pt, anchor=east] (out) at (5.55,2.95) {``good''\\vs\\``poor''};
    \draw[arr] (l5.east) -- (out.west);
    \draw[-{Stealth[length=3pt]}, hotc, thick] (v2.north) -- (v2.north |- 0,3.6);
    \draw[-{Stealth[length=3pt]}, black!70, thick] (out.north) -- (out.north |- 0,3.6);
    \node[lab, anchor=south, text=hotc!90!black] at (v2.north |- 0,3.6) {linear probe};
    \node[lab, anchor=south] at (out.north |- 0,3.6) {answer};
    \node[lab, anchor=east] at (1.4,2.03) {probe};
    \node[lab, anchor=east] at (1.4,1.6) {answer};
    \fill[hotc!80, rounded corners=1pt] (1.5,1.89) rectangle ({1.5+0.903*4.0},2.17);
    \fill[black!45, rounded corners=1pt] (1.5,1.46) rectangle ({1.5+0.617*4.0},1.74);
    \node[anchor=east, font=\scriptsize\bfseries, text=white, inner sep=2pt] at ({1.5+0.903*4.0},2.03) {0.90};
    \node[anchor=east, font=\scriptsize\bfseries, text=white, inner sep=2pt] at ({1.5+0.617*4.0},1.6) {0.62};
    \draw[black!60, densely dotted] ({1.5+0.617*4.0},1.74) -- ({1.5+0.617*4.0},1.89);
    \draw[decorate, decoration={brace, amplitude=3pt, mirror}, thick, black!70] ({1.5+0.617*4.0},1.38) -- ({1.5+0.903*4.0},1.38)
      node[midway, below=3pt, lab] {read-out gap};
    \node[take] at (2.825,0.5) {\textbf{decodable, but not said}\\[1pt] KonIQ SRCC, Qwen3.5-0.8B, one held-out prompt};

    \draw[black!12] (6.4,1.594) -- (11.1,1.594); \node[lab, text=black!50, anchor=east] at (6.4,1.594) {0.6};
    \draw[black!12] (6.4,2.081) -- (11.1,2.081); \node[lab, text=black!50, anchor=east] at (6.4,2.081) {0.7};
    \draw[black!12] (6.4,2.569) -- (11.1,2.569); \node[lab, text=black!50, anchor=east] at (6.4,2.569) {0.8};
    \draw[black!12] (6.4,3.056) -- (11.1,3.056); \node[lab, text=black!50, anchor=east] at (6.4,3.056) {0.9};
    \node[lab, text=black!60, anchor=south west] at (6.0,3.320) {KonIQ SRCC};
    \draw[visc, thick] (6.550,2.145) -- (6.677,2.696) -- (6.805,2.871) -- (6.932,2.956) -- (7.059,3.031) -- (7.186,3.072) -- (7.314,3.022) -- (7.441,2.960) -- (7.568,2.888) -- (7.695,2.752) -- (7.823,2.689) -- (7.950,2.635);
    \draw[lmc!70!black, thick] (8.200,2.582) -- (8.289,2.584) -- (8.378,2.645) -- (8.467,2.689) -- (8.557,2.686) -- (8.646,2.697) -- (8.735,2.656) -- (8.824,2.673) -- (8.913,2.745) -- (9.002,2.725) -- (9.091,2.714) -- (9.180,2.753) -- (9.270,2.783) -- (9.359,2.785) -- (9.448,2.777) -- (9.537,2.790) -- (9.626,2.821) -- (9.715,2.813) -- (9.804,2.820) -- (9.893,2.840) -- (9.983,2.826) -- (10.072,2.820) -- (10.161,2.818) -- (10.250,2.810);
    \draw[black!35, densely dotted] (10.250,2.810) -- (10.800,1.677);
    \fill[black!70] (10.800,1.677) circle (2.2pt);
    \fill[editc] (10.800,2.670) circle (2.2pt);
    \draw[editc, -{Stealth[length=3pt]}, thick] (10.800,1.777) -- (10.800,2.570);
    \node[lab, anchor=north] at (7.250,1.300) {vision blocks};
    \node[lab, anchor=north] at (9.225,1.300) {language layers};
    \node[lab, anchor=north] at (10.800,1.300) {answer};
    \node[lab, anchor=east, text=black!70] at (10.700,1.677) {0.62};
    \node[lab, anchor=east, text=editc!70!black] at (10.700,2.670) {+ME 0.82};
    \node[lab, anchor=south, text=visc!80!black] at (7.186,3.122) {0.90};
    \node[lab, anchor=south, text=lmc!40!black] at (9.804,2.860) {0.85};
    \node[take] at (8.675,0.5) {\textbf{lost after the last layer, in the read-out}\\[1pt] probes (lines) vs.\ answer (dot), Qwen3.5-0.8B};

    \draw[thick, black!60] (12.1,2.05) -- (15.75,2.05);
    \node[lab, text=black!60, anchor=south] at (12.3,2.07) {$h$};
    \foreach \k in {0,...,6} {
      \pgfmathsetmacro{\x}{12.6+\k*0.32}
      \ifnum\k=3
        \node[circle, draw=editc!60!black, fill=editc!70, minimum size=0.24cm, inner sep=0pt] (n\k) at (\x,3.1) {};
      \else
        \node[circle, draw=black!40, fill=white, minimum size=0.24cm, inner sep=0pt] (n\k) at (\x,3.1) {};
      \fi
    }
    \node[lab, text=black!60, anchor=west] at (14.72,3.1) {last MLP};
    \node[lab, text=editc!60!black, anchor=south] at (n3.north |- 0,3.3) {neuron $k$};
    \draw[editc!60!black, thick, -{Stealth[length=3pt]}] (n3.south |- 0,2.05) ++(-0.45,0) -- (n3.south west) node[midway, left, lab] {$u_k$};
    \draw[editc!60!black, thick, -{Stealth[length=3pt]}] (n3.south east) -- (14.01,2.05) node[midway, right, lab] {$d_k$};
    \node[draw=black!50, fill=white, rounded corners=2pt, align=center, font=\scriptsize, inner sep=2pt, minimum width=0.85cm, anchor=west] (gp) at (15.85,2.05) {good\\$-$poor};
    \draw[arr] (15.75,2.05) -- (gp.west);
    \node[lab, text=black!70] at (14.525,1.45) {$u_k$: ridge fit to MOS over 12 prompts;\ \ $d_k \parallel d_{\text{eff}}$};
    \node[take] at (14.525,0.5) {\textbf{2--8k weights, closed form, no training}\\[1pt] 4 VLMs, 12 unseen prompts: KonIQ $+$0.10 to $+$0.22};
    \draw[-{Stealth[length=6pt]}, very thick, black!40] (5.67,2.15) -- (5.83,2.15);
    \draw[-{Stealth[length=6pt]}, very thick, black!40] (11.52,2.15) -- (11.68,2.15);
  \end{tikzpicture}
  \vspace{-2pt}
  \captionof{figure}{%
    \textbf{Lost in read-out.}
    (a) Quality is decodable from a VLM's vision encoder far better than its good/poor answer reveals.
    (b) Ridge probes along Qwen3.5-0.8B (KonIQ test, one held-out prompt).
    (c) Rewriting one last-layer MLP neuron in closed form restores most of it under held-out prompts (\cref{sec:method}); the four VLMs are those of \cref{tab:main}.
  }
  \label{fig:teaser}
\end{strip}

%% file: sec/0_abstract.tex
\begin{abstract}
Vision-language models (VLMs) are widely used as zero-shot image quality assessors, yet their good-vs-poor answers agree only moderately with human opinion.
We ask whether these models fail to see quality or fail to say it.
Layer-wise probes on five VLMs point to the second: quality is linearly decodable from the vision encoder and still from the last language layer, far better than the answer reveals, and about a hundred labels recover most of it.
The loss lies mainly in the final read-out, which partly reads something else: relative to human opinion, it over-rewards colorfulness and under-rewards sharpness on all five models.
We test this by intervention: if the read-out is at fault, a closed-form edit of a single last-layer neuron, fitted to many prompts at once, should close the gap and no more.
Both hold. Under prompts never used for fitting, the edit recovers 86--100\% of what a ridge head on the last-layer state, fitted under the same prompts, finds beyond the answer on KonIQ. Where that gap is small, it gains little, too little for our transfer criterion on the two largest models; where the head itself is below the answer, it lowers the score; and it works only when written from the middle of the language model on.
Unlike unconstrained fine-tuning, the edit leaves perception question answering intact. It does not carry over to yes-or-no or rating questions, and it reverses the good--poor contrast after sentiment-laden sentences.
Code will be made available.
\end{abstract}

%% file: sec/1_intro.tex
\begin{figure*}[t]
  \centering
  \includegraphics[width=\textwidth]{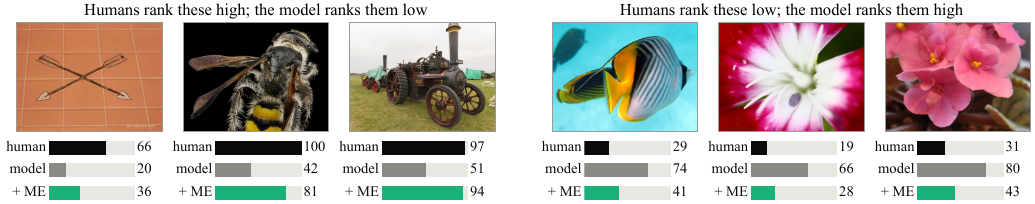}
  \caption{\textbf{Quality decodable from the model but missing from its answer} (Qwen3.5-0.8B, KonIQ test).
    Percentile ranks among the 2,015 test images by human opinion, by the model's good$-$poor answer and after ME (means over 12 held-out prompts).
    Of the 182 images the model misplaces by at least 40 percentiles, three per direction are drawn at random; ME brings 61 of them within 15 percentiles.}
  \label{fig:examples}
\end{figure*}

\section{Introduction}
\label{sec:intro}

Vision-language models (VLMs) are increasingly used as zero-shot image quality assessors: one asks about the quality of an image and compares the next-token logits of ``good'' and ``poor''~\cite{wu2024qbench}.
This needs no labels, but the scores agree only moderately with human opinion. Across the five VLMs we study, the Spearman correlation (SRCC) with mean opinion scores (MOS) ranges from 0.50 to 0.81 on four standard datasets, while models fine-tuned for the task exceed 0.90~\cite{wu2024qalign,you2025deqa}.
The usual remedy is to train on quality data~\cite{wu2024qinstruct,wu2024qalign,you2025deqa,li2025qinsight}.
We first ask a simpler question: does the model fail to see image quality, or does it see quality and fail to say it?

The answer is the second.
Linear probes along the model (\cref{fig:teaser}a,b) decode quality from the middle of the vision encoder about as well as fine-tuned models score (KonIQ SRCC 0.90--0.92), and nearly as well from the last language layer (0.84--0.91), while the answer reaches 0.54--0.80.
The main drop comes at the last step, where the final state is turned into the words ``good'' and ``poor''; we call this step the \emph{read-out}.
This holds on all five models. About a hundred labels recover most of what 2,000 give (0.8B, 4B), and on 0.8B the same labels on a random encoder never reach the answer: the labels tell the read-out where to look rather than what to see.
The read-out does not merely add noise; it partly reads something else.
Compared with human opinion, it over-rewards colorfulness and under-rewards sharpness on all five models, and on four of them it favors images that an aesthetic predictor rates highly.
\cref{fig:examples} illustrates this: vivid images that people rate low are placed high, and plainer, well-captured ones are placed low.

This points to a direct test. Model editing changes what a language model says by rewriting a few MLP weights~\cite{meng2022rome,dai2022neurons}; if the loss lies in the read-out, such an edit of the read-out alone should close the gap, by as much as a ridge head on the last-layer state, fitted under the same prompts, beats the answer, and no more.
We test this by rewriting one neuron of the last MLP layer in closed form, fitted to many prompts at once (\cref{fig:teaser}c); we call it the multi-prompt edit (ME).
The prediction holds in both directions.
Under prompts never used for fitting, the edited model scores within 0.031 SRCC of such a head on five models and four datasets, and it gains most where the gap is large (KonIQ: $+$0.10 to $+$0.22 on four of them).
Where the gap is small, as for Qwen3.5-9B and 27B on LIVE Challenge and BID, it gains little; where the head is below the answer, as on synthetic distortions for the larger Qwen models, the edit is too.
Written into earlier layers, it works from the middle of the language model on, but not before.
The closed form matches gradient tuning of the same weights, and its gains transfer without refitting to two new test sets (six of eight model--dataset pairs).

The edit leaves accuracy on perception question answering (LLVisionQA~\cite{wu2024qbench}) within 0.3 points and every factual answer unchanged. It is not free of side effects: the rewritten neuron acts at every token and shifts ``good'' and ``poor'' in unrelated text (\cref{sec:limits}).
Q-Align-style training~\cite{wu2024qalign} scores higher on KonIQ under its own protocol, but without a constraint it gives up perception question answering, and fully fine-tuned models no longer answer simple factual questions; a KL constraint avoids this, at the cost of a training run.
The edit is also specific to what it was fitted on: it must be refit for AI-generated images~\cite{li2023agiqa}, and it does not carry over to a yes-or-no question.
Our contributions are:
\begin{itemize}
  \item \textbf{Where quality is lost.} In five VLMs from two families, quality is linearly decodable along the model far better than the answer reveals, and it is lost mainly in the final read-out, which over-rewards colorfulness and under-rewards sharpness.
  \item \textbf{A closed-form read-out edit.} Rewriting one last-layer neuron (2,048--10,240 weights, solved in about 2\,s from cached states, no training) substantially improves the zero-shot answer on authentic distortions under held-out prompts, transfers without refitting to new test sets, and matches gradient tuning of the same weights.
  \item \textbf{The gap as a prediction.} The edit recovers 86--100\% of what a ridge head on the last-layer state finds beyond the answer on KonIQ, and where that head is no better than the answer, neither is the edit; the head--answer gap thus tells in advance what repairing the read-out can gain.
  \item \textbf{Perception question answering kept.} The edit leaves perception question answering and factual answers intact, whereas unconstrained quality fine-tuning loses 4--44 points of LLVisionQA accuracy.
\end{itemize}

%% file: sec/2_related.tex
\section{Related Work}
\label{sec:related}

\paragraph{VLMs for image quality assessment.}
Q-Bench~\cite{wu2024qbench} established the zero-shot protocol we build on: ask a VLM about the quality of an image and compare the probabilities of ``good'' and ``poor''.
Earlier, CLIP-IQA~\cite{wang2023clipiqa} scored quality zero-shot by its similarity to antonym text prompts, and LIQE~\cite{zhang2023liqe} trained this vision--language correspondence on quality labels.
Most later work closes the gap to human opinion by training. Q-Instruct~\cite{wu2024qinstruct} tunes on low-level descriptions, Q-Align~\cite{wu2024qalign} on discrete quality levels, Compare2Score~\cite{zhu2024compare2score} on pairwise comparisons and DeQA-Score~\cite{you2025deqa} on score distributions; DepictQA~\cite{you2024depictqa} and Co-Instruct~\cite{wu2024coinstruct} produce descriptions and comparisons, and Q-Insight~\cite{li2025qinsight} and VisualQuality-R1~\cite{wu2025visualqualityr1} train quality reasoning with reinforcement learning.
These update millions to billions of parameters and rarely report effects on other abilities; Q-Debias~\cite{chen2026qdebias} instead corrects perception bias at inference time.
Our edit changes 2,048--10,240 weights in closed form and leaves the query unchanged, but needs labels; we compare it with Q-Align-style full and LoRA~\cite{hu2022lora} training (\cref{sec:train}).
AI-generated and graphics images have human ratings~\cite{li2023agiqa,li2024aigiqa20k,zhang2023cgiqa,li2026r4cgqa} and preference models~\cite{xu2023imagereward,kirstain2023pickapic}, and multimodal judges are fine-tuned~\cite{wang2024maagiqa,zhang2025qeval} or used zero-shot, which A-Bench~\cite{zhang2025abench} finds unreliable.

\paragraph{Model editing and latent knowledge.}
Locate-then-edit methods rewrite a stored fact in closed form: ROME~\cite{meng2022rome} applies a rank-one update to one MLP layer, MEMIT~\cite{meng2023memit} scales this to thousands of facts, and AlphaEdit~\cite{fang2025alphaedit} projects updates so that preserved knowledge is protected.
Other work attributes facts to neurons~\cite{dai2022neurons}, learns editors~\cite{mitchell2022mend}, adds task vectors~\cite{ilharco2023taskarith} or steers activations~\cite{li2023iti,zou2023repe}; MMEdit~\cite{cheng2023mmedit} and VLKEB~\cite{huang2024vlkeb} benchmark multimodal editing, and edits can degrade general abilities~\cite{gu2024harms}.
Ours is also a closed-form MLP edit, but its target is a continuous perceptual read-out rather than a discrete fact.
Like MEMIT, it fits many inputs at once, here prompts rather than facts, and like AlphaEdit it can confine its update to the null space of inputs it should leave unchanged (\cref{sec:method}).
Editing benchmarks test whether unrelated inputs change~\cite{meng2022rome,cheng2023mmedit}; we measure this for our edit on unrelated text (\cref{sec:limits}).
Linear probes~\cite{alain2016probes} show that language models encode more than they say~\cite{burns2023latent,orgad2025llmsknow}, and in VLMs, vision-centric information that can be decoded from the encoder is often lost in the answer~\cite{fu2025hidden}. We show this for image quality, find where it is lost and act on it.

%% file: sec/3_gap.tex
\section{Where Quality Is Lost}
\label{sec:gap}
We locate the loss with probes; the label curve and the layer sweep also use the edit of \cref{sec:method}.

\paragraph{Scoring.}
Following Q-Bench~\cite{wu2024qbench}, the zero-shot quality score of image $i$ under prompt $p$ is
\begin{equation}
  s(i,p) = \ell_{\text{good}}(i,p) - \ell_{\text{poor}}(i,p),
  \label{eq:score}
\end{equation}
where $\ell$ are the next-token logits after a fixed answer prefix (\eg, ``The quality of the image is''), with thinking disabled.
We use two sets of 12 prompts that share no question or prefix: a \emph{fit} set, the only prompts any edit sees, and a \emph{held-out} set, used only for testing.
Unless stated otherwise, we report the Spearman correlation (SRCC) with MOS, averaged over the 12 held-out prompts, with 95\% bootstrap confidence intervals that resample images and prompts (2,000 draws; on KADID, grouped by reference image).

\paragraph{Data.}
Unless stated otherwise, labels come from a pool of 2,000 KonIQ-10k~\cite{hosu2020koniq} training images; 1,000 KonIQ validation images serve only for reporting and for choosing checkpoints of trained baselines.
We test on one synthetic and three authentic datasets: the KonIQ test split (2,015 images), a KADID-10k~\cite{lin2019kadid} subset with one random level of each of 25 distortions per reference (2,025 images), LIVE Challenge~\cite{ghadiyaram2016livec} (1,169) and BID~\cite{ciancio2011bid} (586).

\paragraph{Layer-wise probes.}
We fit ridge probes from internal states to MOS on the 2,000 pool images, separately for every layer and for three kinds of state: the patch mean of each vision-encoder block, the image-token mean of each language-model layer, and the last token of each language-model layer, where the answer is produced.
The regularization and the best probe of each model are chosen by 5-fold cross-validation on the pool alone.
We analyze the three Qwen3.5 models under two prompts, one from each set. Before seeing any score, we fixed what counts as a gap: the best probe must beat the answer by at least 0.03, with its CI above zero, on at least three of the four datasets.

\begin{figure}[t]
  \centering
  \includegraphics[width=\linewidth]{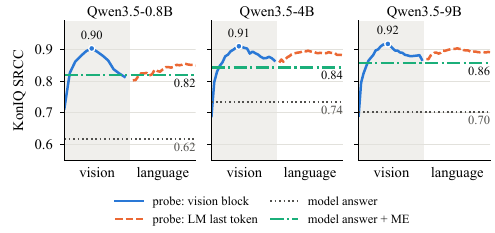}
  \caption{\textbf{Where quality is decodable.}
    KonIQ test SRCC of ridge probes on each vision block and language layer, against the answer before and after ME (one held-out prompt).}
  \label{fig:probes}
\end{figure}

\paragraph{The states carry more than the answer.}
On all three models, the probe selected on the pool sits in a mid-level vision block (block 5 of 12 on 0.8B, 10 of 24 on 4B, 12 of 27 on 9B). On the KonIQ test set it reaches 0.90--0.92, the last-layer probe 0.84--0.89, and the model's answer to the same prompt only 0.62--0.80 (\cref{fig:probes}).
The gap meets our criterion in five of the six model--prompt pairs; in the sixth (4B, held-out prompt), LIVE Challenge misses it by 0.0002.
Gemma 4 E2B and Qwen3.5-27B show the same pattern on KonIQ under the same held-out prompt: the best vision block reaches 0.91 on both (block 10 of 16 and 12 of 27), the last-layer probe 0.85 and 0.91, and the answer 0.54 and 0.79 (both probes minus the answer: CIs above zero).
On KADID, however, the KonIQ-fitted probes are no better than the answer, and on 4B and 9B they are worse (\cref{sec:limits}).

\paragraph{The information comes from visual features.}
On KonIQ, the same probe reaches 0.88 on an ImageNet-only ViT-B/16, close to Qwen3.5-0.8B's own encoder (0.90). It reaches 0.67 on BRISQUE features and 0.56 on a randomly initialized copy of 0.8B's encoder, a control task~\cite{hewitt2019control} that shows what labels and dimensions give by themselves.
Quality is thus carried largely by visual features of the kind an ImageNet-trained encoder learns.

\begin{figure}[t]
  \centering
  \includegraphics[width=\linewidth]{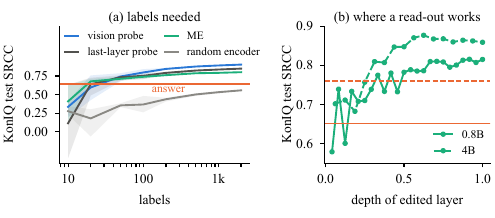}
  \caption{\textbf{How little the read-out needs.}
    (a) KonIQ test SRCC against the number of pool labels (0.8B; mean and range over five subsets; ME: 12 held-out prompts, probes: one).
    (b) The read-out written into one language layer at a time (0.8B solid, 4B dashed, base horizontal; 500 KonIQ test images, real forward passes).}
  \label{fig:localize}
\end{figure}

\paragraph{How many labels reveal it.}
Fitting on random subsets of the pool (five per size, with regularization chosen inside each subset) shows how little supervision the read-out needs (\cref{fig:localize}a).
On 0.8B, 100 labels let the vision probe close 59\% of the gap between the answer (0.64) and the full-pool probe (0.90), and ME reaches 0.74, 61\% of its full gain; with only 20 labels, ME already beats the answer.
On 4B (answer 0.74), ME gets half of its full gain with 50 labels and 74\% with 100.
As the random encoder stays below the answer even with 2,000 labels, the labels point the read-out at information the model already has.
Without labels, no direction we tried reaches the answer (0.62; first principal component 0.32--0.47, sharp-minus-blurry descriptions 0.57).

\paragraph{The loss is mainly in the read-out.}
From the vision encoder to the last language layer, SRCC on KonIQ drops only a little (0.90 to 0.84--0.85 on 0.8B, 0.91 to 0.88--0.89 on 4B). From the last-layer state to the answer it drops much further (to 0.62--0.65 and 0.74--0.80), as it does on Gemma and 27B.
An intervention confirms the location. Written into one language layer at a time, the single-neuron read-out of \cref{sec:method} (fitted to the 12 fit prompts) works as well as in the last layer from layer 16 of 24 on 0.8B (except layer 19) and from layer 14 of 32 on 4B (paired CIs with the last-layer edit include zero), and worse in earlier layers (\cref{fig:localize}b).
From mid-depth on, it only needs to be read out.

\paragraph{What the answer reads instead.}
The answer and a probe on the last-layer state agree only in part: under the 12 held-out prompts, their outputs correlate at 0.67--0.86 on KonIQ (Spearman; 0.44--0.61 after controlling for MOS).
To see where they differ, we correlate both scores with image statistics given MOS (partial Spearman; supplement). This analysis is exploratory; the only hypothesis we fixed in advance concerned the aesthetic score.
Compared with human opinion, the answer over-rewards colorfulness and under-rewards sharpness. On all five models and on both KonIQ and LIVE Challenge, its partial correlation with colorfulness lies above the probe's and that with sharpness below it (paired CIs exclude zero), even though colorfulness is unrelated to MOS on KonIQ ($-$0.02) and sharpness is, among these statistics, the one most correlated with MOS (0.60).
On four of five models, the answer also over-rewards the LAION aesthetic predictor~\cite{schuhmann2022aesthetic,schuhmann2022laion} (KonIQ: 0.29--0.36 vs.\ the probe's 0.11--0.22; not on 0.8B), as \cref{fig:examples} suggests.
After ME, all of these correlations move to within 0.07 of the probe's.
In part, then, the answer reads how appealing an image looks rather than how well it was captured.

%% file: sec/5_method.tex
\begin{table*}[t]
  \centering
  \footnotesize
  \setlength{\tabcolsep}{4pt}
  \caption{\textbf{The model's own good$-$poor answer under 12 held-out prompts} (SRCC; PLCC after a 4-parameter logistic mapping).
    Base is zero-shot; ME and MTC (gradient tuning of the same weights): 2,000 KonIQ pool labels, means over five versions; $^\dagger$/$^\ddagger$refit on the AGIQA-3K / R4-CGQA pool (\cref{sec:exp:agiqa}).
    KonIQ--BID were consulted in development, AGIQA-3K and R4-CGQA added after the recipe was fixed.
    LIQE was trained on KonIQ-10k; its KonIQ values (gray) may include its training images.}
  \label{tab:main}
  \begin{tabular}{llcccccccccccc}
    \toprule
    & & \multicolumn{2}{c}{KonIQ~\cite{hosu2020koniq}} & \multicolumn{2}{c}{KADID~\cite{lin2019kadid}} & \multicolumn{2}{c}{LIVE-C~\cite{ghadiyaram2016livec}} & \multicolumn{2}{c}{BID~\cite{ciancio2011bid}} & \multicolumn{2}{c}{AGIQA-3K~\cite{li2023agiqa}} & \multicolumn{2}{c}{R4-CGQA~\cite{li2026r4cgqa}} \\
    \cmidrule(lr){3-4} \cmidrule(lr){5-6} \cmidrule(lr){7-8} \cmidrule(lr){9-10} \cmidrule(lr){11-12} \cmidrule(lr){13-14}
    Model & Method & SRCC & PLCC & SRCC & PLCC & SRCC & PLCC & SRCC & PLCC & SRCC & PLCC & SRCC & PLCC \\
    \midrule
    \multicolumn{2}{l}{CLIP-IQA~\cite{wang2023clipiqa} (zero-shot)} & 0.642 & 0.725 & 0.513 & 0.535 & 0.702 & 0.720 & 0.590 & 0.567 & 0.649 & 0.702 & 0.214 & 0.251 \\
    \multicolumn{2}{l}{LIQE~\cite{zhang2023liqe} (trained on KonIQ)} & \textcolor{gray}{0.897} & \textcolor{gray}{0.905} & 0.722 & 0.722 & 0.864 & 0.866 & 0.705 & 0.695 & 0.725 & 0.792 & 0.128 & 0.216 \\
    \multicolumn{2}{l}{CLIP + ridge (2,000 labels)} & 0.794 & 0.825 & 0.640 & 0.648 & 0.753 & 0.790 & 0.779 & 0.810 & 0.853$^\dagger$ & 0.906 & 0.469$^\ddagger$ & 0.539 \\
    \midrule
    \multirow{3}{*}{Qwen3.5-0.8B} & Base & 0.642 & 0.727 & 0.546 & 0.591 & 0.694 & 0.761 & 0.662 & 0.715 & 0.687 & 0.772 & 0.109 & 0.172 \\
     & +MTC & 0.773 & 0.831 & 0.578 & 0.612 & 0.774 & 0.825 & 0.755 & 0.787 & 0.800$^\dagger$ & 0.858 & 0.306$^\ddagger$ & 0.343 \\
     & +ME & \textbf{0.800} & \textbf{0.847} & \textbf{0.593} & \textbf{0.623} & \textbf{0.781} & \textbf{0.828} & \textbf{0.763} & \textbf{0.793} & \textbf{0.825}$^\dagger$ & \textbf{0.876} & \textbf{0.360}$^\ddagger$ & \textbf{0.385} \\
    \midrule
    \multirow{3}{*}{Qwen3.5-4B} & Base & 0.742 & 0.799 & \textbf{0.698} & \textbf{0.704} & 0.779 & 0.825 & 0.799 & 0.827 & 0.750 & 0.830 & 0.184 & 0.243 \\
     & +MTC & 0.822 & 0.867 & 0.663 & 0.682 & \textbf{0.813} & \textbf{0.861} & 0.832 & 0.856 & 0.836$^\dagger$ & 0.883 & 0.377$^\ddagger$ & 0.435 \\
     & +ME & \textbf{0.842} & \textbf{0.882} & 0.640 & 0.664 & 0.810 & \textbf{0.861} & \textbf{0.839} & \textbf{0.861} & \textbf{0.848}$^\dagger$ & \textbf{0.893} & \textbf{0.416}$^\ddagger$ & \textbf{0.475} \\
    \midrule
    \multirow{3}{*}{Gemma 4 E2B} & Base & 0.570 & 0.637 & 0.496 & 0.522 & 0.585 & 0.620 & 0.599 & 0.635 & 0.623 & 0.743 & 0.153 & 0.180 \\
     & +MTC & 0.783 & 0.822 & 0.535 & 0.568 & 0.722 & 0.763 & 0.726 & 0.752 & 0.804$^\dagger$ & 0.861 & 0.342$^\ddagger$ & 0.396 \\
     & +ME & \textbf{0.791} & \textbf{0.833} & \textbf{0.572} & \textbf{0.596} & \textbf{0.736} & \textbf{0.777} & \textbf{0.742} & \textbf{0.769} & \textbf{0.818}$^\dagger$ & \textbf{0.869} & \textbf{0.368}$^\ddagger$ & \textbf{0.414} \\
    \midrule
    \multirow{3}{*}{Qwen3.5-9B} & Base & 0.709 & 0.780 & \textbf{0.686} & \textbf{0.700} & 0.794 & 0.832 & 0.792 & 0.825 & 0.718 & 0.812 & 0.148 & 0.278 \\
     & +MTC & 0.819 & 0.866 & 0.656 & 0.671 & 0.807 & 0.860 & \textbf{0.834} & \textbf{0.864} & 0.837$^\dagger$ & 0.889 & 0.369$^\ddagger$ & 0.435 \\
     & +ME & \textbf{0.834} & \textbf{0.875} & 0.632 & 0.653 & \textbf{0.816} & \textbf{0.864} & \textbf{0.834} & 0.861 & \textbf{0.842}$^\dagger$ & \textbf{0.892} & \textbf{0.382}$^\ddagger$ & \textbf{0.464} \\
    \bottomrule
  \end{tabular}
\end{table*}

\section{A Closed-form Read-out Edit}
\label{sec:method}

If the loss lies in the final read-out, rewriting only that read-out should recover what a ridge head on the last-layer state finds beyond the answer, under prompts never used for fitting.
The \emph{multi-prompt edit} (ME) does this: it replaces one neuron of the last MLP layer with a linear read-out of quality, fitted to 12 prompts in closed form.

\paragraph{Notation.}
Let $h$ be the last-token residual state entering the final RMSNorm, with weight $g$, and let $x$ be the input of the last MLP layer at that token (it depends on image $i$ and prompt $p$), whose output is $\sum_k a_k(x)\,(u_k^\top x)\,d_k$ with gate $a_k(x)=\mathrm{act}(w_k^\top x)$, up-projection row $u_k$ and down-projection column $d_k$.
To first order, with $\mathrm{rms}(h)$ held at its mean $\overline{\mathrm{rms}}(h)$ over the pool, adding $\Delta h$ to the residual changes the score in \cref{eq:score} by $d_{\text{eff}}^\top \Delta h$, with
\begin{equation}
  d_{\text{eff}} = \frac{(e_{\text{good}} - e_{\text{poor}}) \odot (1+g)}{\overline{\mathrm{rms}}(h)},
  \label{eq:deff}
\end{equation}
where $e$ are rows of the output (unembedding) matrix.
Neuron $k$ thus contributes $a_k(x)\,(u_k^\top x)\,d_{\text{eff}}^\top d_k$ to the score.

\paragraph{Which neuron.}
We choose the neuron without labels.
Among the half of last-layer neurons with the smallest $\|u_k\|\,\|d_k\|$, we keep those whose gate $a_k$ varies by at most 10\% (standard deviation over mean) across pool images and fit prompts, and rank them by their gate floor, the smallest mean gate over the fit prompts, highest first.
The first makes the neuron cheap to overwrite; the second keeps its gate nearly constant, so that a new read-out acts almost linearly on $x$ for every image and prompt, and it is the one that matters: of 40 random neurons fitted the same way on four models, the 5 whose gate varies by at most 10\% match ME within 0.01 on KonIQ, while all 17 that vary by over 50\% lose 0.03--0.23.
The five top-ranked neurons give five versions, ME$_1$ to ME$_5$. Their scores are nearly identical, so we report them as one result with its range rather than as independent replications.

\paragraph{Closed-form fit.}
Let $s_p(i)$ be the base score of image $i$ under fit prompt $p$, $\sigma_p$ its standard deviation over the pool, and $z(i)$ the standardized MOS.
We want the edited model's score to move toward $\gamma\,\sigma_p\,z(i)$, a target on the scale of each prompt's own scores.
Removing neuron $k$ leaves $r_p(i) = s_p(i) - a_k(x)\,(u_k^\top x)\,d_{\text{eff}}^\top d_k$, and a new neuron with read-out vector $c$ adds $a_k(x)\,c^\top x$.
We solve the ridge problem
\begin{equation}
  \min_{c}\; \sum_{p}\sum_{i}\Big(a_k(x)\,c^\top x - \big[\gamma\,\sigma_p\,z(i) - r_p(i)\big]\Big)^2 + \lambda\|c\|^2,
  \label{eq:ridge}
\end{equation}
over the 2,000 pool images and the 12 fit prompts, with features and targets centered within each prompt, since SRCC ignores offsets.
We then remove from $c$ its component along the mean input $\bar{x}$, so that $c^\top\bar{x} = 0$ and, as its gate is nearly constant, the new neuron adds almost nothing to the score on average over pool images and fit prompts.
A single prompt's mean score can still move (on 0.8B by $-$7.6 to $+$9.8 logits), which matters where answers depend on a threshold (\cref{sec:limits}).
Finally, we write $c$ into the weights:
\begin{equation}
  u_k \leftarrow c/\beta, \qquad d_k \leftarrow \beta\, d_{\text{eff}} / \|d_{\text{eff}}\|^2, \qquad \beta = \|c\| / \|u_k\|,
\end{equation}
which keeps the norm of $u_k$ unchanged and makes the neuron's contribution to the score equal to $a_k(x)\,c^\top x$.
The edit changes $2d$ weights for hidden size $d$: 2,048 on Qwen3.5-0.8B, 8,192 on 9B and 10,240 on 27B.

\paragraph{Choosing $\lambda$ and $\gamma$ without the test prompts.}
We hold out 400 pool images and split the 12 fit prompts into four folds of three.
For each fold, we fit \cref{eq:ridge} on the other nine prompts and 1,600 images and score the 400 held-out ones under the fold's three prompts.
The pair $(\lambda,\gamma)$ with the highest mean SRCC, from $\lambda \in \{10^{-1},\dots,10^{5}\}$ and $\gamma \in \{0.5, 1, 2\}$, is refitted on all images and prompts.
Every version selected $\gamma=2$, the largest; a wider grid selects larger $\gamma$ but changes test SRCC by less than 0.005 (supplement). Held-out prompts and test sets are never used.

\paragraph{A text-side constraint.}
Because the neuron's gate is nearly constant, the new read-out acts on every input, including text without images.
Restricting $c$ to the null space of text-only states keeps the solution in closed form, as in null-space editing~\cite{fang2025alphaedit}.
We take the last-token inputs $x_t$ of the last MLP for 130 text-only sentences (50 topics with a positive and a negative review, and 30 neutral facts; none is an evaluation sentence) and project the features of \cref{eq:ridge} and its solution onto the orthogonal complement of $\{x_t\}$ and the pool mean $\bar{x}$, which replaces the mean removal above; then $c^\top x_t = 0$ exactly.
We call this edit NME and report its effect in \cref{sec:limits}. A penalty $\mu\sum_t(a_k(x_t)\,c^\top x_t)^2$ over every token of WikiText-2 instead trades SRCC against the use of ``good'' and ``poor'' in text (supplement).

\paragraph{Cost and exactness.}
Fitting needs one forward pass per pool image and fit prompt, and solving takes about 2\,s; the linearization in \cref{eq:deff} only builds the target.
Since only the final norm and the unembedding follow the edited layer, the edited model's scores can be computed exactly from cached states; before testing, we checked them against real forward passes of the edited model (Pearson $\geq 0.99$ under every held-out prompt and model).

\paragraph{Other architectures.}
For the untied 9B model, $e$ comes from the output layer. Gemma~4 adds a normalization and per-layer inputs after the MLP, so there $d_{\text{eff}}$ is the label-free mean gradient of the score with respect to the MLP output.

\paragraph{Origin of the structure.}
We began by letting LLM agents, which have automated parts of research~\cite{lu2024aiscientist} and interpretability experiments~\cite{shaham2024maia}, search for weight edits under fixed protocols (19 runs; supplement). The single-neuron structure comes from the first five (Claude Opus 5.5), which, given the weights of Qwen3.5-0.8B, at most 50 logged forward passes and a locked test set, wrote a label-fitted read-out into the last layer to raise SRCC under one prompt; a scripted layer sweep under the same budget finds an equally good site.
Fitted to that one prompt, their edits did not carry over to others. With the same neurons and solver, an edit fitted to their prompt alone (SE) reaches 0.83 on that prompt but only 0.72 on average over the 12 held-out prompts, whereas ME reaches 0.80 on both and beats the base model under every held-out prompt.

%% file: sec/6_experiments.tex
\section{Experiments}
\label{sec:exp}

\subsection{Setup}
\label{sec:exp:setup}
\paragraph{Models and data.}
We edit Qwen3.5-0.8B, 4B and 9B~\cite{qwen2026qwen35} and Gemma~4 E2B~\cite{gemmateam2026gemma4} with the same recipe and 2,000 KonIQ pool images (\cref{sec:method}).
9B and Qwen3.5-27B ran on a second machine, so no paired statistic compares 9B with another model. 27B is edited with ME only, from its FP8 weights, since the bfloat16 weights do not fit in 48\,GB; it is not in \cref{tab:main}, and we did not evaluate its side effects or run it on SPAQ and FLIVE.
Test sets, prompts and scores follow \cref{sec:gap}.
Before testing, we fixed a transfer criterion: a gain of at least 0.03, CI above zero, on three of four datasets.

\paragraph{Baselines.}
\emph{MTC} (multi-prompt tuning with a mean constraint) tunes the same two vectors as ME, with the same pool, prompts and neurons, by gradient descent on a correlation loss under a constraint that keeps the mean score unchanged; its hyperparameters were selected on the pool with the same budget as ME's.
\emph{Q-Align-style training} fine-tunes the whole model on KonIQ with the Q-Align recipe~\cite{wu2024qalign} (full fine-tuning for 0.8B and LoRA~\cite{hu2022lora} with rank 64 for 4B; three seeds; either the 2,000 pool labels or the 7,058 KonIQ training labels).
We also evaluate the public \emph{Q-ReAlign} models~\cite{qfuture2026qrealign} (Mini, Lite and Pro, built on Qwen3.5-0.8B, 4B and 9B), trained with the Q-Align recipe on KonIQ, SPAQ, KADID, AVA and LSVQ.
Outside the VLM, \cref{tab:main} lists zero-shot CLIP-IQA~\cite{wang2023clipiqa}, LIQE~\cite{zhang2023liqe} trained on KonIQ, and ridge regression on frozen CLIP ViT-L/14 embeddings with ME's 2,000 labels.

\subsection{Main results}
\label{sec:exp:main}

\cref{tab:main} shows that ME improves all four models on the three authentic-distortion datasets.
The gain is largest where the base model is weakest (KonIQ: $+$0.22 on Gemma, $+$0.10--0.13 on the larger Qwen models).
ME meets the transfer criterion on 0.8B, 4B and Gemma (five of five versions each), but not on 9B, where the LIVE Challenge interval reaches zero, or on Qwen3.5-27B, where only KonIQ improves significantly ($+$0.07).
On KADID the edit helps 0.8B and Gemma slightly (lower confidence bounds $\geq$0.004). On the two larger Qwen models it lowers SRCC, by 0.05--0.06 on average over prompts and by up to 0.21 under single prompts (supplement); we return to this in \cref{sec:limits}.
PLCC gives the same conclusions in 23 of 24 pairs.

\paragraph{New test sets.}
We also test on two sets not used before: 2,000 SPAQ images~\cite{fang2020spaq} disjoint from the SPAQ pool of \cref{tab:ablation}, and the 1,827-image FLIVE test split~\cite{ying2020paq2piq}. The KonIQ-fitted edit meets the transfer criterion with all five versions in six of eight model--dataset pairs: all four on FLIVE (0.43 to 0.47--0.48 on 0.8B, 0.39 to 0.48 on 4B and 9B, 0.32 to 0.45--0.46 on Gemma) and 0.8B and Gemma on SPAQ, where 4B and 9B stay at 0.88 and 0.87.

\begin{figure}[t]
  \centering
  \includegraphics[width=\linewidth]{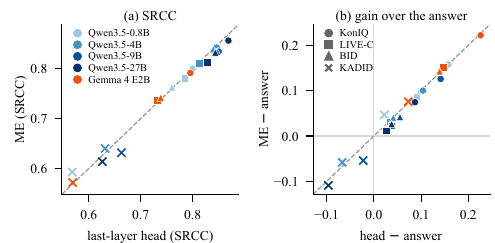}
  \caption{\textbf{The edit reaches the last-layer head.}
    (a) SRCC of a ridge head on the state entering the final norm, read outside the model, against ME (same pool and prompts); (b) the same as gains over the base answer. Dashed: $y=x$.}
  \label{fig:meprobe}
\end{figure}

\paragraph{The edit reaches the last-layer head.}
If the loss lies in the read-out, ME should score as well as a ridge head on the model's own last-layer state, fitted with the same pool, prompts and selection but read outside the model.
This is not a given: ME acts inside the model, through the final normalization, and under prompts it never saw.
On 0.8B, heads on the last MLP's input and on the state entering the final norm score 0.803 and 0.800 on KonIQ, against ME's 0.800 (paired CIs include zero; \cref{tab:ablation}). Across five models and four datasets, ME differs from the latter head by at most 0.031 (mean difference 0.01; least squares $y = 0.95x + 0.035$, Pearson $r = 0.99$; \cref{fig:meprobe}a).
ME thus recovers 96--100\% of the head's KonIQ gain over the answer on 0.8B, 4B and Gemma and 86--89\% on 9B and 27B (\cref{fig:meprobe}b), and where the head falls below the answer (KADID on 4B, 9B and 27B), the edit does too.
Heads on layer 16, or fitted to one prompt, score 0.02--0.03 and 0.05--0.07 lower. The probes of \cref{sec:gap} score higher (0.8B: 0.849) because they are fitted and tested under one prompt.
When choosing among AI-generated images, ridge probes on the same states pick as well as the refit ME (supplement).
As a metric, ME does not exceed such a head. Its value is that it locates the gap and, because it changes the model, tests what no outside probe can: the read-out still works when written from mid-depth on (\cref{fig:localize}b), other antonym pairs improve with it (\cref{sec:exp:formats}), and on 0.8B and Gemma open descriptions become more accurate (\cref{sec:limits}).
The CLIP baseline matches ME on KonIQ for 0.8B and Gemma, trails it on 4B and 9B, and beats it on AGIQA-3K and R4-CGQA (\cref{tab:main}).

\paragraph{How large the read-out gap is.}
The gap between the head and the answer does not depend on ME, and it varies across models. On KonIQ it is 0.23 on Gemma, 0.14--0.16 on 0.8B and 9B, and 0.09--0.10 on 4B and 27B. On LIVE Challenge and BID it is 0.09--0.15 on 0.8B and Gemma but only 0.03--0.06 on the three larger Qwen models, so the gains there (0.01--0.04) do not all clear the transfer criterion.
ME recovers what a last-layer read-out can (the head: KonIQ 0.80--0.87), not the further loss between the vision encoder and the last layer (\cref{sec:gap}).

\paragraph{Closed form vs.\ gradient tuning.}
On the same weights, ME and MTC are close (\cref{tab:main}).
Fitted on KonIQ, neither is better on three of the four datasets for any model, so at this site the closed form matches gradient tuning. Per dataset, ME is higher on KonIQ (0.8B, 4B) and MTC on KADID (4B, 9B).
Refit on AGIQA-3K and R4-CGQA, MTC is up to 0.07 below ME (significantly for 16 of 40 versions, and never above).
Extending MTC's grid where its choice lay at an end never made it significantly better than ME.
Training-free alternatives do not help either: Q-Debias~\cite{chen2026qdebias} gives no reliable gain on the authentic datasets, a steering vector changes 0.8B's SRCC by at most $+$0.008, and few-shot prompting lowers it (supplement).

\input{sec/6b_training}

\subsection{Ablations}
\label{sec:exp:ablation}
\begin{table}[t]
  \centering
  \footnotesize
  \setlength{\tabcolsep}{3.5pt}
  \caption{\textbf{Ablations on Qwen3.5-0.8B} (SRCC, 12 held-out prompts, mean of five versions).
    Bold: best per column, $^*$excluded; $^*$KADID column (Base 0.552, ME 0.602) on 40 held-out reference images.}
  \label{tab:ablation}
  \begin{tabular}{lcccc}
    \toprule
    Variant & KonIQ & KADID & LIVE-C & BID \\
    \midrule
    Base & 0.642 & 0.546 & 0.694 & 0.662 \\
    ME (ours) & 0.800 & \textbf{0.593} & 0.781 & \textbf{0.763} \\
    \midrule
    Unconstrained gradient tuning & 0.792 & 0.562 & 0.777 & 0.759 \\
    Rescale 300 neurons, no fit & 0.685 & 0.586 & 0.723 & 0.700 \\
    \midrule
    \multicolumn{5}{l}{\emph{Ridge head read outside the model}} \\
    \quad on layer-16 state & 0.772 & 0.563 & 0.765 & 0.742 \\
    \quad on last-layer state & \textbf{0.803} & 0.572 & \textbf{0.787} & 0.761 \\
    \quad \ldots{} added to the answer & 0.801 & 0.586 & 0.780 & \textbf{0.763} \\
    \quad \ldots{} fitted to one prompt & 0.731 & 0.510 & 0.713 & 0.713 \\
    \midrule
    Pool: SPAQ & 0.736 & 0.582 & 0.759 & 0.734 \\
    Pool: KADID & 0.657 & 0.670$^*$ & 0.648 & 0.636 \\
    Pool: KonIQ + KADID & 0.754 & 0.689$^*$ & 0.751 & 0.734 \\
    \bottomrule
  \end{tabular}
\end{table}

\cref{tab:ablation} varies one part of the recipe at a time. Fitting to 12 prompts matters most (\cref{sec:method}): with one prompt, gradient tuning keeps almost none of the gain (KonIQ 0.663). Fitting one neuron beats rescaling 300 existing ones (by 0.06--0.12 on the authentic datasets). The pool decides which distortions improve: a mix of KonIQ and KADID lifts all four datasets but costs 0.03--0.05 on the authentic ones.

\subsection{AI-generated and computer-graphics images}
\label{sec:exp:agiqa}
We repeat the study on AGIQA-3K~\cite{li2023agiqa}, split by text prompt into a pool (1,988 images) and a test set (994 images).
The KonIQ-fitted ME does not help here (+0.001 to +0.005, not significant), but the same recipe refit on the AGIQA pool raises SRCC by 0.10--0.20 on all four models (\cref{tab:main}).
On computer graphics (R4-CGQA~\cite{li2026r4cgqa}, 220 test images), base models are near chance (0.11--0.18) and the KonIQ edit does not transfer. Refit on 2,000 graphics images, ME reaches 0.36--0.42 (+0.22 to +0.25, CIs above zero); Q-ReAlign reaches 0.23--0.30 (supplement).

\subsection{Other answer formats}
\label{sec:exp:formats}
Without refitting, the KonIQ-fitted edit also improves other antonym pairs read at the same position (``high''/``low'', ``excellent''/``bad'', ``sharp''/``blurry''; KonIQ $+$0.04 to $+$0.06 on 0.8B and $+$0.008 to $+$0.03 on 4B, all CIs above zero).
The Q-Align five-level expectation improves on 4B ($+$0.10) but not reliably on 0.8B.
The gain does not carry over to a different question: the yes$-$no score for ``Is this a high-quality image?'' and a 1--5 rating change by less than 0.01 on both models.
So the edit aligns the good$-$poor read-out and its close neighbors, not every way of asking.

%% file: sec/6b_training.tex
\subsection{Comparison with training the whole model}
\label{sec:train}

\begin{figure}[t]
  \centering
  \includegraphics[width=\linewidth]{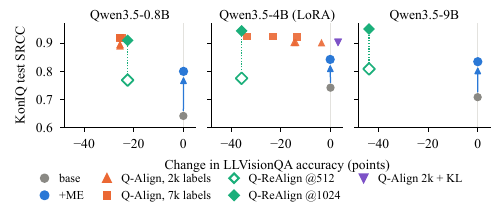}
  \caption{\textbf{Two ways to improve quality answers.}
    KonIQ test SRCC against the change in LLVisionQA accuracy. Base and ME: 12 held-out prompts; trained models: their own Q-Align score, one marker per seed; KL: LoRA with a KL penalty.}
  \label{fig:tradeoff}
\end{figure}

Under its own scoring, training beats ME on KonIQ by 0.06--0.12, and also on LIVE Challenge, which no trained model saw (\cref{fig:tradeoff}). Under our held-out prompts, both 0.8B models we trained keep this lead, while the 4B LoRA models and Q-ReAlign fall below ME.
Every model trained without a constraint loses 4--44 points of LLVisionQA accuracy, and the fully trained ones answer none of 40 factual questions, giving a sentence about image quality instead (Q-ReAlign Pro: ``the capital of the image is good''). ME changes accuracy by at most 0.3 points and keeps every factual answer.
A standard constraint avoids this: with a KL penalty to the base model's answers on generic questions, 4B LoRA keeps its KonIQ score, gains 2.9--3.2 LLVisionQA points and keeps the factual answers (three seeds; held-out 0.838--0.859, ME 0.842). Training therefore need not lose other abilities; ME gets there without training, and LoRA stopped early gets there too (supplement).
Fitting ME takes about 2\,s once pool states are cached (24,000 forward passes), a 4B LoRA run 16--21\,min on a 48\,GB GPU.

%% file: sec/7_costs.tex
\section{Effects and Limitations}
\label{sec:limits}

\paragraph{Effects beyond quality assessment.}
ME leaves LLVisionQA accuracy and factual answers essentially unchanged (\cref{tab:side}; within 0.3 points also for the AGIQA and graphics refits).
Because its gate is nearly constant, however, the edited neuron acts at every token.
It weakens the contrast between ``good'' and ``poor'' after positive and negative sentences, both after ten short reviews (Valence) and after 872 SST-2 validation sentences continued with ``In one word, I found it'' (SST val.); after SST-2 it reverses the sign on all three models. On 0.8B it also raises WikiText-2 perplexity by 1.5--3.8\% and lowers SST-2 accuracy by 2.5--3.7 points; on 4B and 9B, perplexity by 0.5--1.3\% (supplement).

\paragraph{A text-side constraint helps only in part.}
NME (\cref{sec:method}) restores the ten-review gap (on Gemma only partly) at almost no cost in SRCC, and on 4B and 9B even exceeds the base model's gap, since it does not control the removal of neuron $k$'s original output.
Its 130 sentences, however, share the format of those ten reviews. After the SST-2 sentences it keeps 87\% of the base gap on 0.8B but only 21--24\% on 4B and 9B, and on 0.8B SST-2 accuracy falls further than with ME.
A constraint built from generic text does no better. Projecting out the top-$r$ principal directions of 4,800 WikiText token states (NME-G, as in AlphaEdit~\cite{fang2025alphaedit}) keeps KonIQ within 0.003 of ME, but the SST-2 gap stays at 0.29 of the base 1.75 on 0.8B, $-$1.12 of 2.25 on 4B and $-$0.53 of 1.58 on 9B (ME: $-$0.62, $-$1.97, $-$0.57), and 0.8B's SST-2 accuracy stays at 81.0\% (base 86.4\% in the same run). NME-G thus fails our criterion, fixed beforehand, of keeping 90\% of the base SST-2 gap on all three models (supplement), and \cref{tab:main} reports ME.
Under NME, LLVisionQA stays within 0.6 points, and under both constraints factual answers change by at most one of 40.
An edit in the vision encoder, which text-only inputs never reach, can avoid these effects. One agent's single-neuron edit of 0.8B's vision block~9 keeps the valence and SST-2 gaps, at KonIQ SRCC 0.759 (ME$_1$: 0.801; supplement).

\begin{table}[t]
  \centering
  \footnotesize
  \setlength{\tabcolsep}{1.6pt}
    \caption{\textbf{Effects of ME and of its text-constrained version NME} (brackets: range over versions). KonIQ: SRCC; LLVQA: LLVisionQA accuracy change (points); Valence and SST val.: ``good''$-$``poor'' log-probability after positive minus after negative sentences (ten reviews; 872 SST-2 sentences, measured in a later run, not on Gemma); SST-2: accuracy (\%).}
  \label{tab:side}
  \resizebox{\linewidth}{!}{%
  \begin{tabular}{llccccc}
    \toprule
    & & KonIQ & LLVQA $\Delta$ & Valence & SST val. & SST-2 \\
    \midrule
    \multirow{3}{*}{0.8B} & base & 0.642 & -- & 6.5 & 1.75 & 86.7 \\
     & ME & 0.800 & [$-$0.2, $+$0.3] & [$-$0.1, 0.4] & [$-$0.77, $-$0.56] & [83.0, 84.2] \\
     & NME & [0.800, 0.801] & [$-$0.5, $-$0.3] & [6.2, 6.3] & [1.49, 1.55] & [80.2, 80.8] \\
    \midrule
    \multirow{3}{*}{4B} & base & 0.742 & -- & 7.4 & 2.25 & 93.8 \\
     & ME & 0.842 & [$-$0.3, 0.0] & [3.3, 3.6] & [$-$2.11, $-$1.83] & [93.8, 93.9] \\
     & NME & [0.842, 0.844] & [$-$0.3, $+$0.2] & [8.2, 8.3] & [0.44, 0.58] & [93.8, 94.0] \\
    \midrule
    \multirow{3}{*}{Gemma} & base & 0.570 & -- & 5.3 & -- & 87.2 \\
     & ME & 0.791 & [$-$0.3, $-$0.1] & [2.7, 2.8] & -- & 87.3 \\
     & NME & [0.786, 0.791] & [$-$0.2, $-$0.1] & [3.5, 3.8] & -- & [87.0, 87.2] \\
    \midrule
    \multirow{3}{*}{9B} & base & 0.709 & -- & 8.2 & 1.58 & 94.8 \\
     & ME & 0.834 & [$-$0.1, $+$0.3] & [5.5, 5.8] & [$-$0.68, $-$0.48] & [95.1, 95.2] \\
     & NME & [0.832, 0.837] & [$-$0.1, $+$0.3] & [8.6, 8.7] & [0.30, 0.37] & [94.4, 94.6] \\
    \bottomrule
  \end{tabular}}
\end{table}

\paragraph{Generated answers.}
Generated text follows the edited score only in part. Single-word answers barely change, and on 0.8B completed sentences become more positive, because the centering holds over the 12 fit prompts but not for each prompt separately.
Scored with a lexicon of quality words, open descriptions improve on 0.8B and Gemma (0.8B, BID: 0.47 to 0.58) but not on 4B, and 5.5--13\% of Gemma's descriptions degenerate into a repeated ``poor''.
Asked which of two images is better, the edited model stays within 1.5 points of the base (supplement).

\paragraph{Scope of the evidence.}
All five models come from two families, Qwen3.5 and Gemma~4, and all 12 held-out prompts end where an adjective is expected.
On KADID, where the KonIQ-fitted probes are no better than the answer, ME lowers SRCC on 4B and 9B (\cref{sec:exp:main}) and, as for generated images, must be refit.
We chose the structure after seeing which agent edit transferred best, and we consulted KonIQ--BID during development, so these columns of \cref{tab:main} test the recipe on new models, not on new data; SPAQ and FLIVE do (\cref{sec:exp:main}).

%% file: sec/8_conclusion.tex
\section{Conclusion}
\label{sec:conclusion}

We asked whether VLMs fail to see image quality or fail to say it.
They see it: quality, decodable from ordinary visual features, survives through the language model and is lost mainly in the read-out that turns the last state into ``good'' or ``poor'' and favors colorful over sharp images.
Rewriting that read-out alone, one neuron in closed form, recovers what a ridge head on the last-layer state, fitted under the same prompts, finds beyond the answer (86--100\% on KonIQ), and no more, so that gap indicates what repairing the read-out can gain.
Unlike unconstrained whole-model training, the edit keeps perception question answering. Its costs: it shifts ``good'' and ``poor'' in unrelated text, which text-side constraints remove only in part, and must be refit per kind of distortion. We see ME less as a new metric than as a test of where a model's knowledge is lost.

\par

%% file: sec/9_llm_statement.tex
\section*{LLM Usage Statement}
{\parfillskip=0pt plus 0.6\columnwidth
We used Claude Opus 5.5 (Anthropic), run through Claude Code, in four parts of this work.
(i)~Method: 19 agent runs explored weight edits under fixed protocols; the single-neuron structure of ME comes from the first five (\cref{sec:method}, supplement).
(ii)~Code and experiments: the model wrote and ran most of the experiment, analysis and plotting code under the authors' direction; the protocols were fixed before testing and logged.
(iii)~Writing: the model searched the related literature and drafted and revised the paper and the supplement, including editing and grammar.
(iv)~Ideation: the model helped the authors analyze their ideas; the research question, the main ideas and the contributions are the authors' own.
The authors take full responsibility for the content of this paper.\par}

%% file: sec/X_suppl.tex
\maketitlesupplementary

\section{Implementation and Reproducibility}
\label{sec:suppl:impl}

\subsection{Prompts}
\label{sec:suppl:prompts}
\Cref{tab:suppl:prompts} lists the 24 prompts.
Every prompt is a user turn with the image followed by the question, and an assistant turn that starts with the answer prefix, with thinking disabled; the score of \cref{eq:score} is read at the next token.
The two sets share no question or prefix.
Fit prompt F1 is the agents' prompt (\cref{sec:agents}); F4--F12 are the nine prompts no agent saw, used to audit the agents' edits; F2 and F3 had been used in an earlier check the agents might have seen.
The layer-wise probes of \cref{sec:gap} use F1 (``fit'') and H1 (``held-out'').
Edits never see H1--H12, which are used only for testing.

\begin{table*}[t]
  \centering
  \scriptsize
  \setlength{\tabcolsep}{4pt}
  \caption{\textbf{The 24 prompts} (question; answer prefix). $^a$The agents' prompt; $^b$the nine audit prompts; $^c$the held-out prompt of the layer-wise probes.}
  \label{tab:suppl:prompts}
  \begin{tabular}{@{}lp{0.47\textwidth}p{0.42\textwidth}@{}}
    \toprule
    & Question & Answer prefix \\
    \midrule
    \multicolumn{3}{@{}l}{\emph{Fit prompts (the only prompts any edit sees)}} \\
    F1$^a$ & Rate the quality of the image. & The quality of the image is \\
    F2 & How would you judge the technical quality of this photo? & The technical quality of this photo is \\
    F3 & Assess the perceptual quality of this image, considering blur, noise, exposure and compression. & The perceptual quality of this image is \\
    F4$^b$ & Describe the visual quality of this image. & The visual quality of this image is \\
    F5$^b$ & What is the image quality like? & The image quality is \\
    F6$^b$ & How good is the quality of this photograph? & The quality of this photograph is \\
    F7$^b$ & Give your opinion on how this picture looks technically. & Technically, this picture looks \\
    F8$^b$ & Would a photographer consider this a high quality photo? & A photographer would consider the quality of this photo \\
    F9$^b$ & Assess this image as an image quality expert. & As an expert, I rate the image quality as \\
    F10$^b$ & Is the picture quality acceptable for publishing? & For publishing, the picture quality is \\
    F11$^b$ & Rate this photo's overall quality. & Overall, the quality of this photo is \\
    F12$^b$ & How does this image look to a human viewer? & To a human viewer, the image quality is \\
    \midrule
    \multicolumn{3}{@{}l}{\emph{Held-out prompts (testing only)}} \\
    H1$^c$ & What is your verdict on the quality of this picture? & My verdict is that the quality of this picture is \\
    H2 & Comment on the quality of the photo. & The photo's quality is \\
    H3 & Evaluate how well this image was captured. & The capture quality of this image is \\
    H4 & What do you think of the overall look of this image? & I think the overall quality of this image is \\
    H5 & Is this image any good from a technical standpoint? & From a technical standpoint, the image quality is \\
    H6 & A customer asks whether this photo's quality is satisfactory. What do you say? & I would say the quality of this photo is \\
    H7 & Please grade the picture quality of the image below. & The picture quality of the image is \\
    H8 & Summarize the perceived quality of this image in one word. & In one word, the perceived quality is \\
    H9 & How would a viewer describe the quality of this shot? & A viewer would describe the quality of this shot as \\
    H10 & Judge this image only on its visual quality, not its content. & Ignoring the content, the visual quality is \\
    H11 & Was this photo taken well? & In terms of quality, this photo is \\
    H12 & Tell me about the fidelity of this photograph. & The fidelity of this photograph is \\
    \bottomrule
  \end{tabular}
\end{table*}

\subsection{Models, data and hardware}
\label{sec:suppl:setup}
\paragraph{Models.}
Qwen3.5-0.8B, 4B and 9B, Qwen3.5-27B (official FP8 release) and Gemma~4 E2B, all instruction-tuned checkpoints, run in bfloat16 with the chat template of each model and thinking disabled.
For 27B, every FP8 weight is dequantized to bfloat16 before its matrix product, since the fused FP8 kernel needs a newer PyTorch than our server's, and the last-layer MLP is stored in bfloat16 so that the edit can be written exactly.
Qwen3.5-0.8B, 4B and Gemma tie the output matrix to the input embedding; 9B and 27B do not.

\paragraph{Images.}
Images go through each model's own processor at their stored resolution: KonIQ-10k in its 512$\times$384 release, KADID-10k at 512$\times$384, LIVE Challenge at 500$\times$500 and AGIQA-3K as released.
BID, SPAQ, FLIVE and R4-CGQA images are larger, and we first resize their long side to 512 pixels (Lanczos, aspect ratio kept).
The resolution paragraph of \cref{sec:suppl:train} uses the 1,024$\times$768 KonIQ release instead.
The KADID subset takes, for each of the 81 reference images, one random level of each of the 25 distortions (2,025 images); its confidence intervals resample reference images.

\paragraph{Hardware.}
One NVIDIA RTX 4070 Ti SUPER (16\,GB) and one NVIDIA RTX 4090 with 48\,GB.
Most runs of 0.8B, 4B and Gemma used the former; 9B and 27B used the latter, so no statistic pairs them with the other models.

\subsection{The read-out direction}
\label{sec:suppl:deff}
In Qwen3.5 the logits are $\ell = E\,\mathrm{norm}(h)$ with $\mathrm{norm}(h) = (1+g)\odot h/\mathrm{rms}(h)$ and $\mathrm{rms}(h) = (\tfrac{1}{d}\|h\|^2+\epsilon)^{1/2}$, so the score of \cref{eq:score} is $s = w^\top h/\mathrm{rms}(h)$ with $w = (e_{\text{good}} - e_{\text{poor}})\odot(1+g)$.
Its gradient with respect to the residual state is
\begin{equation}
  \nabla_h s = \frac{1}{\mathrm{rms}(h)}\Big(w - \frac{w^\top \hat{h}}{d}\,\hat{h}\Big), \qquad \hat{h} = h/\mathrm{rms}(h),
\end{equation}
where $\|\hat{h}\|^2 = d$ (ignoring $\epsilon$).
The second term is the part of $w$ along $\hat{h}$; relative to the first it has size $|\cos(w, h)|$.
ME drops it and replaces $\mathrm{rms}(h)$ by its mean over the pool, which gives \cref{eq:deff}: one direction for all images and prompts.
The approximation only builds the regression target and the direction in which the new neuron writes; every score is computed exactly through the final norm (\cref{sec:method}).
As a check, the cosine between $d_{\text{eff}}$ and the mean of the exact gradient over pool images under F1 is 0.9999 on 4B, 0.99998 on 9B and 0.9999 on 27B.
For 9B and 27B, $E$ is the separate output matrix.
Writing $d_k \leftarrow \beta\,d_{\text{eff}}/\|d_{\text{eff}}\|^2$ makes $d_{\text{eff}}^\top d_k = \beta$, so the neuron adds $a_k(x)\,(u_k^\top x)\,\beta = a_k(x)\,c^\top x$ to the score to first order.

On Gemma~4, the MLP output passes through a post-feed-forward norm, a per-layer input block and a layer scalar before the final norm, and the logits are soft-capped.
There, $d_{\text{eff}}$ is defined as the mean gradient of the score with respect to the MLP output over the pool images under F1, computed without labels; everything else is unchanged, and scores are computed through Gemma's full remaining computation, including the soft cap.
This was fixed before Gemma was run, so Gemma tests the recipe with one stated change.

\subsection{Hyperparameters and baselines}
\label{sec:suppl:hyper}
\paragraph{Ridge probes and heads.}
Penalty $\lambda \in \{10^{-1}, 1, \dots, 10^{5}\}$, chosen by 5-fold cross-validation on the pool.
The layer-wise probes of \cref{sec:gap} are fitted and tested under one prompt (F1 or H1).
The heads of \cref{sec:exp:main} (``last-layer head'') are fitted, like ME, on the pool under the 12 fit prompts (features centered per prompt) with ME's folds, and tested under the 12 held-out prompts; they are the heads of \cref{fig:meprobe} and \cref{tab:suppl:gapgain}.
The same state thus gives 0.849 for the single-prompt probe and 0.800 for the head on 0.8B KonIQ: the head must transfer to prompts it was not fitted on.

\paragraph{ME.}
$\lambda \in \{10^{-1}, \dots, 10^{5}\}$, $\gamma \in \{0.5, 1, 2\}$ (extended grid in \cref{sec:suppl:gamma}), selected as in \cref{sec:method}; the five neurons are the five top-ranked by the rule of \cref{sec:method}.

\paragraph{MTC.}
The same neurons, prompts and pool as ME; the up- and down-projection vectors of neuron $k$ are tuned with Adam to minimize the squared difference between the per-prompt standardized score and the standardized MOS, plus a weight-decay term (decay $\in \{0, 10^{-2}, 10^{-1}, 1\}$, relative to the original norms) and a penalty on the change of each prompt's mean score (weight $\in \{10, 10^2, 10^3\}$).
Learning rate $\in \{10^{-4}, 3\cdot10^{-4}, 10^{-3}, 3\cdot10^{-3}\}$, up to 2,000 full-batch steps, the step chosen on held-out pool images every 20 steps, all selected with ME's folds (extended grid in \cref{sec:suppl:full}).

\paragraph{Q-Align-style training.}
Following Q-Align~\cite{wu2024qalign}, the model is trained to answer one of five level words (excellent, good, fair, poor, bad) after ``The quality of the image is'', under nine question templates (``How would you rate the quality of this image?'' and eight paraphrases), with MOS binned into five equal-width levels; its own score is the expectation over the five level words with weights 1, 0.75, 0.5, 0.25, 0.
0.8B: full fine-tuning, learning rate $2\cdot10^{-5}$; 4B: LoRA~\cite{hu2022lora} with rank 64, $\alpha=128$, dropout 0.05 on the language model, learning rate $2\cdot10^{-4}$; AdamW, batch 16, two epochs, 3\% warm-up, three seeds.
\emph{KL}: the same LoRA run plus a per-token KL penalty to the frozen base model's greedy answers (48 tokens) to four generic questions (``Describe the image.'', ``What objects are in this picture?'', ``What is happening in this image?'', ``Describe the colors and the lighting of this photo.'') on 256 pool images.

\paragraph{Outside baselines.}
The CLIP-IQA row of \cref{tab:main} uses pyiqa's \texttt{clipiqa} metric with default settings on the same image files, without training.
The CLIP + ridge row is a supervised reference with the same labels as ME: frozen CLIP ViT-L/14 image embeddings and a ridge regression whose penalty is chosen by 5-fold cross-validation on the same 2,000-image pool (the AGIQA-3K or R4-CGQA pool for those columns), applied once to each test set.
Like the probes, it is an external regressor and leaves the model's answers unchanged.

\paragraph{Statistics.}
SRCC is computed per held-out prompt and averaged.
Confidence intervals are 95\% percentile intervals of 2,000 bootstrap draws that resample images (KADID: reference images; AGIQA-3K: text prompts) and prompts together (``two-way''), paired across the methods compared.
PLCC follows a four-parameter logistic fit per prompt; a fit that collapses to a step falls back to the raw scores.

\subsection{Evaluation and constraint sentences}
\label{sec:suppl:materials}
\paragraph{Facts (40).}
Asked with ``Answer briefly.'' appended, greedy, 12 tokens; correct if the reply contains the answer: What is the capital of France?; What is the capital of Japan?; What is the capital of Italy?; What is the capital of Spain?; What is the capital of Egypt?; What is the capital of Russia?; What is the capital of China?; What is the capital of Germany?; What color is a clear daytime sky?; What color is fresh grass?; What color is snow?; What color is a ripe banana?; How many days are in a week?; How many months are in a year?; How many legs does a spider have?; How many sides does a triangle have?; How many sides does a hexagon have?; What is 2 + 2?; What is 7 + 5?; What is 9 - 3?; What is 6 * 7?; What is 10 / 2?; What planet do humans live on?; What is the largest planet in the solar system?; What star is at the center of the solar system?; What gas do humans need to breathe to live?; What is the chemical symbol for gold?; What is the chemical symbol for sodium?; What liquid is H2O?; What is the opposite of hot?; What is the opposite of up?; What is the opposite of good?; What is the opposite of high?; What is the plural of mouse?; What is the past tense of go?; Which animal says meow?; Which animal says woof?; Who wrote Romeo and Juliet?; What language is spoken in Brazil?; What is the freezing point of water in Celsius?.

\paragraph{Neutral starts (10).}
Under the instruction ``Complete the sentence.'': ``The capital of France is''; ``Water boils at a temperature of''; ``The first month of the year is''; ``A triangle has three''; ``The largest ocean on Earth is the''; ``The sun rises in the''; ``Cats and dogs are both''; ``The color of grass is''; ``Paris is a city in''; ``Two plus three equals''.

\paragraph{Valence pairs (10).}
Under ``Complete the sentence.''; we compare the log-probabilities of `` good'' and `` poor'' at the next token.
\begin{itemize}\itemsep0pt
  \item The food was delicious and the staff were kind. Overall, the meal was / The food was cold and the staff were rude. Overall, the meal was
  \item The plot was gripping and the acting was superb. Overall, the movie was / The plot was dull and the acting was awful. Overall, the movie was
  \item The room was spotless and the bed was comfortable. Overall, the hotel was / The room was filthy and the bed was broken. Overall, the hotel was
  \item The battery lasts for days and it never crashes. Overall, the phone is / The battery dies within hours and it often crashes. Overall, the phone is
  \item I could not put it down and I loved the ending. Overall, the book is / I could not finish it and I hated the ending. Overall, the book is
  \item The band played beautifully and the crowd was thrilled. Overall, the concert was / The band played badly and the crowd was angry. Overall, the concert was
  \item They fit perfectly and still look new after a year. Overall, the shoes are / They hurt my feet and fell apart in a week. Overall, the shoes are
  \item She explains clearly and always helps her students. Overall, the teacher is / She explains nothing and ignores her students. Overall, the teacher is
  \item The controls are smooth and the story is moving. Overall, the game is / The controls are clumsy and the story is boring. Overall, the game is
  \item It was sunny and warm with a light breeze. Overall, the weather was / It was freezing and stormy with heavy rain. Overall, the weather was
\end{itemize}

\paragraph{SST-2.}
All 872 validation sentences: sentiment accuracy under ``Is the sentiment of this sentence positive or negative?'' (prefix ``The sentiment is'', `` positive'' vs.\ `` negative''), and the SST-2 gap of \cref{sec:limits} under ``Continue the text.'' with ``In one word, I found it'' appended.

\paragraph{NME's 130 constraint sentences.}
For each of 50 topics (car, laptop, flight, museum, park, bakery, coffee, haircut, dentist, bicycle, headphones, sofa, garden, train ride, bus service, app, website, podcast, song, course, lecture, workshop, gym, pizza, wine, hike, beach, festival, wedding, interview, meeting, team, project, report, essay, recipe, jacket, watch, tent, boat trip, zoo, library, apartment, neighborhood, doctor, mechanic, plumber, translation, conference, vacation), one positive and one negative review, built as ``[sentence] Overall, the [topic] was'', with the sentence cycling through five positive (``Everything worked as promised and I would gladly return.'' / ``It exceeded my expectations in every way.'' / ``I enjoyed every minute and would recommend it to anyone.'' / ``It was a pleasure from start to finish.'' / ``The people involved were friendly and very helpful.'') and five negative ones (``Nothing worked as promised and I would never return.'' / ``It fell short of my expectations in every way.'' / ``I regretted every minute and would warn anyone against it.'' / ``It was a disappointment from start to finish.'' / ``The people involved were unfriendly and no help at all.''); and 30 neutral starts: ``The chemical symbol for gold is''; ``A week has''; ``The boiling point of water in Fahrenheit is''; ``The planet closest to the sun is''; ``A hexagon has''; ``The author of Hamlet is''; ``The capital of Japan is''; ``The tallest mountain on Earth is''; ``A year usually has''; ``The main gas in the air we breathe is''; ``The longest river in Africa is''; ``Ice is the solid form of''; ``The square root of nine is''; ``Spiders have''; ``The currency of the United Kingdom is the''; ``Light travels faster than''; ``The opposite of north is''; ``The smallest prime number is''; ``Bees make''; ``The language spoken in Brazil is''; ``A dozen means''; ``The Pacific is an''; ``The Moon orbits the''; ``Ten times ten equals''; ``The hardest natural material is''; ``Penguins live mostly in the''; ``The heart pumps''; ``A bicycle has two''; ``The freezing point of water in Celsius is''; ``Germany is a country in''.
None of them is an evaluation sentence (checked in code).

\section{Locating the Loss}
\label{sec:suppl:local}

\subsection{Probes and answers on every dataset}
\label{sec:suppl:probes}
\Cref{tab:suppl:probes} gives the numbers behind \cref{fig:probes} and the gap criterion of \cref{sec:gap} for all three Qwen models, both prompts and all four datasets.
The best probe is the vision block selected by pool cross-validation (the same block for both prompts); the criterion asks that its SRCC exceed the answer's by at least 0.03 with the CI above zero on at least three of four datasets.
It holds in five of the six model--prompt pairs; for 4B under H1, LIVE Challenge misses it by 0.0002 (gain 0.0298, CI lower bound 0.008).
On KADID, the KonIQ-fitted probes are never significantly better than the answer, and on 4B and 9B significantly worse.

\begin{table*}[t]
  \centering
  \scriptsize
  \setlength{\tabcolsep}{2.6pt}
  \caption{\textbf{Answer vs.\ probes} (KonIQ-fitted ridge probes on the 2,000-image pool, one prompt). Ans.: the model's good$-$poor answer; best: probe on the selected vision block (Blk); last: probe on the last token of the last language layer; CI: 95\% CI of best $-$ answer; $^\checkmark$: meets the 0.03 gap with CI above zero.}
  \label{tab:suppl:probes}
  \begin{tabular}{@{}lcc|cccc|cccc|cccc|cccc@{}}
    \toprule
    & & & \multicolumn{4}{c|}{KonIQ} & \multicolumn{4}{c|}{KADID} & \multicolumn{4}{c|}{LIVE Challenge} & \multicolumn{4}{c}{BID} \\
    Model & Pr. & Blk & Ans. & best & last & CI & Ans. & best & last & CI & Ans. & best & last & CI & Ans. & best & last & CI \\
    \midrule
    0.8B & F1 & 5 & 0.653 & 0.903 & 0.840 & [0.23, 0.28]$^\checkmark$ & 0.551 & 0.563 & 0.608 & [$-$0.03, 0.06] & 0.697 & 0.796 & 0.794 & [0.07, 0.13]$^\checkmark$ & 0.693 & 0.843 & 0.779 & [0.11, 0.19]$^\checkmark$ \\
    0.8B & H1 & 5 & 0.617 & 0.903 & 0.849 & [0.26, 0.31]$^\checkmark$ & 0.588 & 0.563 & 0.570 & [$-$0.07, 0.02] & 0.702 & 0.796 & 0.802 & [0.06, 0.13]$^\checkmark$ & 0.662 & 0.843 & 0.783 & [0.14, 0.23]$^\checkmark$ \\
    4B & F1 & 10 & 0.795 & 0.910 & 0.888 & [0.10, 0.13]$^\checkmark$ & 0.713 & 0.554 & 0.667 & [$-$0.19, $-$0.13] & 0.793 & 0.825 & 0.838 & [0.01, 0.05]$^\checkmark$ & 0.810 & 0.856 & 0.857 & [0.02, 0.08]$^\checkmark$ \\
    4B & H1 & 10 & 0.735 & 0.910 & 0.882 & [0.15, 0.20]$^\checkmark$ & 0.732 & 0.554 & 0.698 & [$-$0.21, $-$0.15] & 0.795 & 0.825 & 0.837 & [0.01, 0.05] & 0.802 & 0.856 & 0.861 & [0.02, 0.09]$^\checkmark$ \\
    9B & F1 & 12 & 0.728 & 0.917 & 0.891 & [0.17, 0.21]$^\checkmark$ & 0.674 & 0.588 & 0.693 & [$-$0.12, $-$0.05] & 0.806 & 0.840 & 0.851 & [0.01, 0.05]$^\checkmark$ & 0.796 & 0.867 & 0.858 & [0.04, 0.10]$^\checkmark$ \\
    9B & H1 & 12 & 0.702 & 0.917 & 0.891 & [0.19, 0.24]$^\checkmark$ & 0.728 & 0.588 & 0.667 & [$-$0.17, $-$0.10] & 0.806 & 0.840 & 0.830 & [0.01, 0.05]$^\checkmark$ & 0.809 & 0.867 & 0.857 & [0.03, 0.09]$^\checkmark$ \\
    \bottomrule
  \end{tabular}
\end{table*}

\subsection{Vision probes on Gemma and 27B}
\begin{figure}[t]
  \centering
  \includegraphics[width=\linewidth]{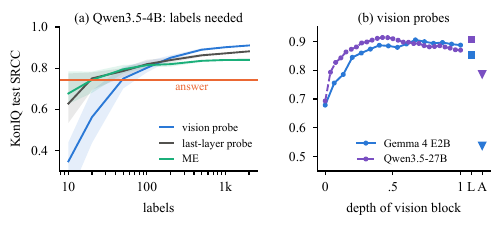}
  \caption{\textbf{(a) Labels needed on 4B} (as \cref{fig:localize}a; mean and range over five subsets per size). \textbf{(b) Vision-block probes on Gemma and 27B} under H1 (KonIQ test SRCC against relative block depth); L: last-layer probe; A: the answer.}
  \label{fig:suppl:curves}
\end{figure}

With the recipe of \cref{fig:probes} (held-out prompt, KonIQ only), the best vision block chosen on the pool is block 10 of 16 on Gemma~4 E2B and block 12 of 27 on Qwen3.5-27B. \Cref{fig:suppl:curves}b shows every block. KonIQ test SRCC: best vision block 0.905 and 0.913, last-layer probe 0.853 and 0.907, answer 0.538 and 0.787; vision minus answer [0.336, 0.398] and [0.110, 0.143], last layer minus answer [0.286, 0.344] and [0.105, 0.134] (95\% CIs).

\subsection{Label efficiency}
\Cref{tab:suppl:labels} gives the curves of \cref{fig:localize}a for both models (\cref{fig:suppl:curves}a for 4B; mean over five pool subsets per size; regularization chosen inside each subset by $K$-fold cross-validation with $K=\min(5,n)$).
A random-encoder cache exists only for 0.8B.

\begin{table}[h]
  \centering
  \footnotesize
  \setlength{\tabcolsep}{3pt}
  \caption{\textbf{KonIQ test SRCC against the number of labels.} Vision: probe on the best vision block; last: probe on the last language layer (one held-out prompt); ME: 12 held-out prompts; rand.: probe on the randomly initialized encoder. Answers: 0.8B 0.642, 4B 0.742.}
  \label{tab:suppl:labels}
  \begin{tabular}{rccccccc}
    \toprule
    & \multicolumn{4}{c}{Qwen3.5-0.8B} & \multicolumn{3}{c}{Qwen3.5-4B} \\
    \cmidrule(lr){2-5} \cmidrule(lr){6-8}
    $n$ & vision & last & ME & rand. & vision & last & ME \\
    \midrule
    10 & 0.335 & 0.115 & 0.409 & 0.277 & 0.347 & 0.628 & 0.677 \\
    20 & 0.596 & 0.638 & 0.678 & 0.181 & 0.562 & 0.750 & 0.743 \\
    50 & 0.743 & 0.734 & 0.712 & 0.356 & 0.749 & 0.786 & 0.794 \\
    100 & 0.797 & 0.752 & 0.739 & 0.367 & 0.801 & 0.819 & 0.814 \\
    200 & 0.845 & 0.792 & 0.761 & 0.438 & 0.850 & 0.839 & 0.819 \\
    500 & 0.877 & 0.823 & 0.787 & 0.502 & 0.890 & 0.862 & 0.836 \\
    1000 & 0.891 & 0.837 & 0.792 & 0.533 & 0.902 & 0.872 & 0.840 \\
    2000 & 0.903 & 0.849 & 0.801 & 0.557 & 0.910 & 0.882 & 0.840 \\
    \bottomrule
  \end{tabular}
\end{table}
\subsection{Label-free directions}
On 0.8B under one held-out prompt (answer 0.617), the first principal component of pool states reaches an absolute SRCC of 0.324 (vision block 5) and 0.466 (last layer), and the difference between last-layer states for six descriptions of sharp and six of blurry photos 0.566.
\subsection{Layer sweep}
\Cref{tab:suppl:sweep} gives the curve of \cref{fig:localize}b. The neuron of each layer is chosen by ME's rule among that layer's neurons, the column points along the last layer's $d_{\text{eff}}$, and $\lambda,\gamma$ are selected on the pool from the linearized score; in the last layer this reproduces ME's neuron (0.8B: 2, 4B: 13).

\begin{table}[h]
  \centering
  \footnotesize
  \setlength{\tabcolsep}{2.5pt}
  \caption{\textbf{Read-out written into one language layer} (500 KonIQ test images, 12 held-out prompts, real forward passes). Base: 0.651 (0.8B), 0.760 (4B). $^\ast$Paired CI against the last-layer edit includes zero or lies above it.}
  \label{tab:suppl:sweep}
  \begin{tabular}{cccccccc}
    \toprule
    \multicolumn{8}{l}{\emph{Qwen3.5-0.8B, layers 1--24}} \\
    0.579 & 0.739 & 0.601 & 0.733 & 0.708 & 0.710 & 0.739 & 0.775 \\
    0.733 & 0.780 & 0.732 & 0.782 & 0.789 & 0.785 & 0.786 & 0.809$^\ast$ \\
    0.810$^\ast$ & 0.808$^\ast$ & 0.795 & 0.800$^\ast$ & 0.813$^\ast$ & 0.815$^\ast$ & 0.807$^\ast$ & 0.815 \\
    \midrule
    \multicolumn{8}{l}{\emph{Qwen3.5-4B, layers 2, 4, \ldots, 32}} \\
    0.702 & 0.711 & 0.682 & 0.757 & 0.809 & 0.807 & 0.847$^\ast$ & 0.847$^\ast$ \\
    0.870$^\ast$ & 0.877$^\ast$ & 0.867$^\ast$ & 0.868$^\ast$ & 0.860$^\ast$ & 0.864$^\ast$ & 0.862$^\ast$ & 0.859 \\
    \bottomrule
  \end{tabular}
\end{table}
\subsection{How the answer and the head relate}
\label{sec:suppl:dirs}
\Cref{tab:suppl:dirs} compares the answer's read-out with the last-layer head (version 1) in two ways.
In weight space, the cosine between $d_{\text{eff}}$ and the head's weights on the same state is 0.03--0.05, at the scale of two random directions ($1/\sqrt{d}$), whereas heads fitted on two disjoint halves of the pool agree at 0.59--0.79.
A small cosine in a space of thousands of dimensions does not make the outputs independent, since the states vary only along a few directions; the outputs are the relevant comparison.
Under the 12 held-out prompts, the two read-outs correlate at 0.67--0.86 on KonIQ and 0.75--0.91 on LIVE Challenge (mean of per-prompt Spearman correlations over test images), and at 0.44--0.61 and 0.59--0.75 after controlling for MOS (partial Spearman correlation, $\rho_{ab\cdot y} = (\rho_{ab}-\rho_{ay}\rho_{by})/\sqrt{(1-\rho_{ay}^2)(1-\rho_{by}^2)}$).
The two read-outs agree in large part and differ in what \cref{sec:suppl:reads} examines.

\begin{table}[h]
  \centering
  \footnotesize
  \setlength{\tabcolsep}{2.2pt}
  \caption{\textbf{Answer read-out vs.\ last-layer head.} cos: cosine of $d_{\text{eff}}$ and the head's weights; halves: cosine of heads fitted on two pool halves; output: Spearman correlation of the two scores, raw / given MOS, mean over held-out prompts.}
  \label{tab:suppl:dirs}
  \begin{tabular}{@{}lrcccc c@{}}
    \toprule
    Model & $d$ & $1/\sqrt{d}$ & cos & halves & KonIQ & LIVE-C \\
    \midrule
    0.8B & 1,024 & 0.031 & 0.046 & 0.77 & 0.75 / 0.51 & 0.84 / 0.67 \\
    4B & 2,560 & 0.020 & 0.053 & 0.77 & 0.81 / 0.54 & 0.90 / 0.75 \\
    Gemma & 1,536 & 0.026 & 0.026 & 0.79 & 0.67 / 0.44 & 0.75 / 0.59 \\
    9B & 4,096 & 0.016 & 0.045 & 0.77 & 0.80 / 0.54 & 0.91 / 0.75 \\
    27B & 5,120 & 0.014 & 0.030 & 0.59 & 0.86 / 0.61 & 0.91 / 0.74 \\
    \bottomrule
  \end{tabular}
\end{table}

\subsection{What the answer reads}
\label{sec:suppl:reads}
\paragraph{Image statistics.}
Brightness is the mean luminance, contrast its standard deviation, colorfulness the measure of Hasler and S\"usstrunk, sharpness the log variance of the Laplacian at a long side of 512 pixels, and the aesthetic score that of the LAION improved aesthetic predictor~\cite{schuhmann2022aesthetic} (an MLP on normalized CLIP ViT-L/14 embeddings, weights sac+logos+ava1-l14-linearMSE). Scores are per-image means of per-prompt z-scores over the 12 held-out prompts; partial Spearman correlations control for MOS through rank residuals, with 2,000 group-bootstrap draws for the paired differences. \Cref{tab:suppl:reads} lists them; the ``probe on the last-layer state'' of \cref{sec:gap} is this head. This analysis is exploratory: the criterion we fixed before computing it concerned only the aesthetic score on KonIQ (larger for the answer than for the last-layer head, and smaller after ME, CIs excluding zero) and holds on 4B, Gemma, 9B and 27B, not on 0.8B; the colorfulness and sharpness differences of \cref{sec:gap} were read from the same table afterwards.

\begin{table*}[t]
  \centering
  \footnotesize
  \setlength{\tabcolsep}{6pt}
  \caption{\textbf{Partial Spearman correlation with image statistics given MOS} (answer / last-layer head / ME). Correlation of each statistic with MOS: KonIQ 0.36, 0.24, $-$0.02, 0.60, 0.40; LIVE Challenge 0.22, 0.11, 0.17, 0.50, 0.50.}
  \label{tab:suppl:reads}
  \begin{tabular}{lccccc}
    \toprule
    & Brightness & Contrast & Colorful. & Sharpness & Aesthetic \\
    \midrule
    \multicolumn{6}{l}{\emph{KonIQ}} \\
0.8B & 0.25 / 0.27 / 0.27 & $-$0.08 / 0.12 / 0.14 & 0.14 / 0.05 / 0.03 & $-$0.22 / 0.13 / 0.16 & 0.16 / 0.18 / 0.13 \\
4B & 0.05 / 0.21 / 0.19 & 0.09 / 0.18 / 0.18 & 0.12 / 0.07 / 0.08 & 0.00 / 0.20 / 0.21 & 0.36 / 0.11 / 0.12 \\
Gemma & 0.09 / 0.20 / 0.19 & $-$0.06 / 0.21 / 0.21 & 0.17 / 0.03 / 0.01 & $-$0.15 / 0.26 / 0.27 & 0.29 / 0.22 / 0.20 \\
9B & 0.06 / 0.19 / 0.18 & $-$0.03 / 0.10 / 0.12 & 0.14 / 0.05 / 0.03 & $-$0.12 / 0.13 / 0.20 & 0.34 / 0.18 / 0.13 \\
27B & 0.11 / 0.20 / 0.18 & 0.10 / 0.16 / 0.16 & 0.11 / 0.05 / 0.04 & 0.04 / 0.17 / 0.18 & 0.31 / 0.16 / 0.14 \\
    \midrule
    \multicolumn{6}{l}{\emph{LIVE Challenge}} \\
0.8B & 0.17 / 0.23 / 0.24 & 0.08 / 0.10 / 0.12 & 0.33 / 0.22 / 0.20 & 0.10 / 0.26 / 0.29 & 0.13 / 0.14 / 0.10 \\
4B & 0.09 / 0.21 / 0.21 & 0.15 / 0.18 / 0.18 & 0.26 / 0.12 / 0.11 & 0.23 / 0.35 / 0.36 & 0.25 / 0.08 / 0.08 \\
Gemma & 0.07 / 0.25 / 0.26 & 0.04 / 0.09 / 0.09 & 0.35 / 0.20 / 0.16 & 0.06 / 0.30 / 0.30 & 0.20 / 0.13 / 0.11 \\
9B & 0.08 / 0.21 / 0.21 & 0.16 / 0.20 / 0.20 & 0.26 / 0.11 / 0.10 & 0.20 / 0.31 / 0.32 & 0.28 / 0.15 / 0.13 \\
27B & 0.15 / 0.28 / 0.28 & 0.15 / 0.15 / 0.13 & 0.27 / 0.12 / 0.11 & 0.24 / 0.31 / 0.31 & 0.25 / 0.07 / 0.02 \\
    \bottomrule
  \end{tabular}
\end{table*}
\section{Edit Results}
\label{sec:suppl:results}

\subsection{Full results with confidence intervals}
\label{sec:suppl:full}
\Cref{tab:suppl:full} adds intervals and per-prompt detail to the KonIQ-fitted rows of \cref{tab:main}.
Our rule for a consistent difference between two methods asks that one be better, with a CI excluding zero, on at least three of the four datasets for at least three of the five paired versions; under this rule neither ME nor MTC is consistently better on any model.
Per dataset, ME is significantly higher than MTC in 26 of 80 version--dataset pairs (all 10 KonIQ pairs of 0.8B and 4B among them) and MTC in 5, all on KADID (4B and 9B), where ME lowers SRCC.
We had predicted that the KonIQ differences would be mostly non-significant; this did not hold.

\paragraph{MTC with an extended grid.}
In 26 of MTC's 60 fits (KonIQ, AGIQA-3K and R4-CGQA pools) the selected learning rate lay at a grid end or training reached the 2,000-step limit.
We added $3\cdot10^{-5}$ or $10^{-2}$ beyond the end reached (with the same four decay values), trained step-limited configurations to 4,000 steps, and kept the original selection rule on the pool.
A new configuration was selected for 14 fits (pool score $+$0.000 to $+$0.025; 8 again at the new grid end), not for the other 12, including 10 of Gemma's 11.
On the test sets, the 14 refits change MTC's SRCC on their own dataset by $-$0.002 to $+$0.014; none is significantly above ME (MTC$-$ME CIs below zero for 0.8B and 4B on KonIQ and 0.8B on AGIQA-3K, including or touching zero otherwise).

\begin{table*}[t]
  \centering
  \footnotesize
  \setlength{\tabcolsep}{4pt}
  \caption{\textbf{KonIQ-fitted ME and MTC on the four test sets} (SRCC, 12 held-out prompts, means over five versions).
    CI: smallest lower and largest upper bound of the two-way 95\% bootstrap CI of ME$-$Base over the five versions.
    Prompts: held-out prompts (of 12) on which ME's mean SRCC exceeds the base model's, and the range of these per-prompt gains.
    MTC$-$ME: range over the five paired versions; Sig.: versions in which ME\,/\,MTC is significantly higher.}
  \label{tab:suppl:full}
  \begin{tabular}{llccccccccc}
    \toprule
    Model & Dataset & Base & ME & ME$-$Base & CI & Prompts & Per-prompt gain & MTC & MTC$-$ME & Sig. \\
    \midrule
    \multirow{4}{*}{0.8B} & KonIQ & 0.642 & 0.800 & $+$0.158 & [$+$0.132, $+$0.184] & 12 & $+$0.112 to $+$0.204 & 0.773 & $-$0.035 to $-$0.024 & 5\,/\,0 \\
     & KADID & 0.546 & 0.593 & $+$0.047 & [$+$0.004, $+$0.087] & 10 & $-$0.047 to $+$0.115 & 0.578 & $-$0.029 to $+$0.000 & 1\,/\,0 \\
     & LIVE-C & 0.694 & 0.781 & $+$0.087 & [$+$0.058, $+$0.117] & 12 & $+$0.035 to $+$0.169 & 0.774 & $-$0.011 to $-$0.004 & 0\,/\,0 \\
     & BID & 0.662 & 0.763 & $+$0.101 & [$+$0.069, $+$0.136] & 12 & $+$0.057 to $+$0.143 & 0.755 & $-$0.011 to $-$0.005 & 0\,/\,0 \\
    \midrule
    \multirow{4}{*}{4B} & KonIQ & 0.742 & 0.842 & $+$0.100 & [$+$0.064, $+$0.131] & 11 & $-$0.018 to $+$0.190 & 0.822 & $-$0.030 to $-$0.011 & 5\,/\,0 \\
     & KADID & 0.698 & 0.640 & $-$0.058 & [$-$0.107, $-$0.015] & 1 & $-$0.206 to $+$0.064 & 0.663 & $+$0.007 to $+$0.047 & 0\,/\,3 \\
     & LIVE-C & 0.779 & 0.810 & $+$0.031 & [$+$0.006, $+$0.058] & 11 & $-$0.037 to $+$0.097 & 0.813 & $-$0.003 to $+$0.010 & 0\,/\,0 \\
     & BID & 0.799 & 0.839 & $+$0.040 & [$+$0.018, $+$0.063] & 11 & $-$0.005 to $+$0.076 & 0.832 & $-$0.008 to $-$0.006 & 1\,/\,0 \\
    \midrule
    \multirow{4}{*}{Gemma} & KonIQ & 0.570 & 0.791 & $+$0.222 & [$+$0.173, $+$0.273] & 12 & $+$0.087 to $+$0.382 & 0.783 & $-$0.014 to $-$0.001 & 2\,/\,0 \\
     & KADID & 0.496 & 0.572 & $+$0.076 & [$+$0.020, $+$0.134] & 11 & $-$0.004 to $+$0.252 & 0.535 & $-$0.060 to $-$0.017 & 4\,/\,0 \\
     & LIVE-C & 0.585 & 0.736 & $+$0.152 & [$+$0.103, $+$0.199] & 12 & $+$0.052 to $+$0.292 & 0.722 & $-$0.023 to $-$0.007 & 2\,/\,0 \\
     & BID & 0.599 & 0.742 & $+$0.143 & [$+$0.098, $+$0.189] & 12 & $+$0.088 to $+$0.218 & 0.726 & $-$0.024 to $-$0.010 & 3\,/\,0 \\
    \midrule
    \multirow{4}{*}{9B} & KonIQ & 0.709 & 0.834 & $+$0.126 & [$+$0.080, $+$0.167] & 12 & $+$0.014 to $+$0.256 & 0.819 & $-$0.021 to $-$0.009 & 1\,/\,0 \\
     & KADID & 0.686 & 0.632 & $-$0.054 & [$-$0.096, $-$0.009] & 3 & $-$0.149 to $+$0.039 & 0.656 & $-$0.007 to $+$0.071 & 0\,/\,2 \\
     & LIVE-C & 0.794 & 0.816 & $+$0.022 & [$-$0.001, $+$0.045] & 11 & $-$0.044 to $+$0.083 & 0.807 & $-$0.017 to $+$0.004 & 2\,/\,0 \\
     & BID & 0.792 & 0.834 & $+$0.042 & [$+$0.014, $+$0.071] & 11 & $-$0.009 to $+$0.111 & 0.834 & $-$0.003 to $+$0.006 & 0\,/\,0 \\
    \bottomrule
  \end{tabular}
\end{table*}
\subsection{Gap and gain}
\Cref{tab:suppl:gapgain} gives the points of \cref{fig:meprobe} as differences from the answer. The last-layer head is that of \cref{sec:exp:main} on the state entering the final norm (fitted on the pool under the 12 fit prompts), and all three scores are means over the 12 held-out prompts and five versions. Over the 20 points, least squares with an intercept gives ME $= 0.948 \times$ head $+ 0.035$ (Pearson $r = 0.993$; mean $|$ME $-$ head$|$ 0.0095, max 0.031), and gain $= 1.02 \times$ gap $- 0.006$ ($r = 0.989$); the second relation is close to the first by construction, since both subtract the same answer.

\begin{table}[h]
  \centering
  \footnotesize
  \setlength{\tabcolsep}{3.5pt}
  \caption{\textbf{Read-out gap / ME gain} (SRCC; gap = last-layer head $-$ answer, gain = ME $-$ answer).}
  \label{tab:suppl:gapgain}
  \resizebox{\linewidth}{!}{%
  \begin{tabular}{lcccc}
    \toprule
    Model & KonIQ & KADID & LIVE-C & BID \\
    \midrule
    0.8B & 0.158 / 0.158 & 0.022 / 0.047 & 0.091 / 0.087 & 0.098 / 0.101 \\
    4B & 0.104 / 0.100 & $-$0.067 / $-$0.058 & 0.034 / 0.031 & 0.041 / 0.040 \\
    Gemma & 0.226 / 0.222 & 0.073 / 0.076 & 0.147 / 0.152 & 0.139 / 0.143 \\
    9B & 0.142 / 0.126 & $-$0.023 / $-$0.054 & 0.038 / 0.022 & 0.055 / 0.042 \\
    27B & 0.087 / 0.075 & $-$0.096 / $-$0.109 & 0.027 / 0.011 & 0.038 / 0.027 \\
    \bottomrule
  \end{tabular}}
\end{table}
\subsection{Extended $\gamma$ grid}
\label{sec:suppl:gamma}
With $\gamma \in \{0.5, 1, 2\}$ every version selected $\gamma=2$, the edge of the grid.
We extended the grid to $\{0.5, 1, 2, 4, 8, 16\}$, selected again by the same pool cross-validation, refitted the versions whose choice changed and scored them on the four test sets (\cref{tab:suppl:gamma}).
The cross-validated SRCC of 4B, Gemma and 9B peaks at $\gamma=4$--8; that of 0.8B still rises at 16, by 0.003 over $\gamma=2$.
The largest test change is $+$0.0043 (9B, KonIQ, where all five paired CIs are above zero); all other changes have CIs that include zero.

\begin{table}[h]
  \centering
  \footnotesize
  \setlength{\tabcolsep}{3pt}
  \caption{\textbf{Extended $\gamma$ grid.} Selected $\gamma$ per version; gain in pool cross-validated SRCC over $\gamma=2$; change in test SRCC of the refitted edit over the original ME (4 datasets, 5 versions, 12 held-out prompts).}
  \label{tab:suppl:gamma}
  \begin{tabular}{lccc}
    \toprule
    Model & Selected $\gamma$ & CV gain & Test $\Delta$ SRCC \\
    \midrule
    0.8B  & 16, 16, 16, 16, 16 & 0.0025--0.0038 & $-$0.0014 to $+$0.0025 \\
    4B    & 4, 4, 4, 2, 4 & 0.0000--0.0005 & $-$0.0010 to $+$0.0016 \\
    Gemma & 8, 4, 4, 4, 4 & 0.0002--0.0013 & $+$0.0001 to $+$0.0021 \\
    9B    & 4, 4, 4, 4, 8 & 0.0005--0.0016 & $-$0.0004 to $+$0.0043 \\
    \bottomrule
  \end{tabular}
\end{table}
\subsection{Random neurons}
\label{sec:suppl:random}
To test the neuron rule of \cref{sec:method}, we replaced ME's neuron by random ones and kept everything else: the same closed-form fit, hyperparameter selection on the pool and test computation.
R-all draws five neurons from all last-layer neurons, R-low five from the low-importance half, dropping only the two gate conditions (all four models).
\Cref{tab:suppl:random} groups the 40 random neurons by how much their gate varies.
Neurons with a stable gate match ME, including two among the 3\% most important neurons, while unstable gates lose up to 0.23.
A random neuron is clearly below ME (ME$-$random CI above zero on at least two authentic datasets) for 1 of 5 versions in each group on 0.8B, but for 3--5 of 5 on 4B and Gemma; on 9B, for 4 of 5 R-all versions and none of R-low.
On 4B, two unstable neurons raise KADID above ME (0.70 vs.\ 0.64) while losing 0.07--0.10 on KonIQ.
The pool selection score flags the failures (\eg, 0.67 and 0.71 on 0.8B vs.\ 0.78--0.80 for the others).

\begin{table}[h]
  \centering
  \footnotesize
  \caption{\textbf{Random neurons} (four models; 40 neurons). KonIQ SRCC of ME minus the random neuron with the same version number, by the variation of the neuron's gate (standard deviation over mean). ME's own neurons vary by 0.02--0.10.}
  \label{tab:suppl:random}
  \begin{tabular}{lcc}
    \toprule
    Gate variation & Neurons & ME $-$ random (KonIQ) \\
    \midrule
    $\leq 0.10$ & 5 & $-$0.008 to $-$0.003 \\
    0.10--0.50 & 18 & $-$0.005 to $+$0.052 \\
    $> 0.50$ & 17 & $+$0.030 to $+$0.226 \\
    \bottomrule
  \end{tabular}
\end{table}
\subsection{Other answer formats}
\Cref{tab:suppl:formats} lists every format for both models (KonIQ test). Antonym pairs and the Q-Align expectation are read at the answer position under the 12 held-out prompts from cached states; the yes/no question and the 1--5 rating need their own prompts and were run with real forward passes for ME$_1$--ME$_3$.

\begin{table}[h]
  \centering
  \footnotesize
  \setlength{\tabcolsep}{3pt}
  \caption{\textbf{Other answer formats} (KonIQ SRCC, base $\to$ range over ME versions; $^\ast$CIs above zero for all versions, $^\circ$CIs below zero, otherwise CIs include zero).}
  \label{tab:suppl:formats}
  \begin{tabular}{lcc}
    \toprule
    Format & 0.8B & 4B \\
    \midrule
    ``high''$-$``low'' & 0.607 $\to$ 0.660--0.663$^\ast$ & 0.706 $\to$ 0.739--0.740$^\ast$ \\
    ``excellent''$-$``bad'' & 0.535 $\to$ 0.579--0.582$^\ast$ & 0.627 $\to$ 0.659--0.661$^\ast$ \\
    ``sharp''$-$``blurry'' & 0.674 $\to$ 0.726--0.731$^\ast$ & 0.700 $\to$ 0.708$^\ast$ \\
    Q-Align five levels & 0.469 $\to$ 0.489--0.513 & 0.691 $\to$ 0.788--0.792$^\ast$ \\
    Yes/no question & 0.679 $\to$ 0.676--0.679 & 0.664 $\to$ 0.664 \\
    1--5 rating & 0.621 $\to$ 0.627--0.628$^\ast$ & 0.731 $\to$ 0.726--0.729$^\circ$ \\
    \bottomrule
  \end{tabular}
\end{table}
\subsection{Comparing two images in one prompt}
\label{sec:suppl:pairwise}
ME is fitted to single-image prompts that read ``good'' against ``poor''.
To test whether it also changes answers to a question it was not fitted to, we drew 500 pairs of KonIQ test images whose MOS differ by at least 10 (of 100) and asked, with both images in one prompt, which has better quality, under three prompts fixed in advance (\eg, ``Which of the two images has better quality, the first or the second?'', answer prefix ``The image with better quality is the'').
We read the logit of `` first'' minus `` second'', asked each pair in both orders and subtracted the two to cancel position bias.
The plan counted ME as improving this question on a model if, for at least 3 of the 5 versions, the paired bootstrap CI (2,000 resamples over pairs) of the accuracy difference to the base model lay above zero.

\begin{table}[h]
  \centering
  \footnotesize
  \setlength{\tabcolsep}{3pt}
  \caption{\textbf{Two-image comparison} (accuracy, mean of three prompts; ME: range over five versions). ``Single image'': each image scored alone under the 12 held-out prompts, then compared.}
  \label{tab:suppl:pairwise}
  \begin{tabular}{lcccc}
    \toprule
    & \multicolumn{2}{c}{Two images, one prompt} & \multicolumn{2}{c}{Single image} \\
    \cmidrule(lr){2-3} \cmidrule(lr){4-5}
    Model & Base & ME & Base & ME \\
    \midrule
    0.8B & 71.5\% & 71.7--73.0\% & 84.8\% & 93.8--94.2\% \\
    4B   & 87.6\% & 87.7--88.0\% & 88.4\% & 95.4--95.6\% \\
    \bottomrule
  \end{tabular}
\end{table}

The criterion is not met on either model: the CI lies above zero for 2 of 5 versions on 0.8B (ME$_1$ $+$0.5 to $+$2.7 points, ME$_4$ $+$0.4 to $+$2.5) and for none on 4B.
Asked in one order only, 0.8B rarely names the first image (20\% of pairs for the base model, 11--13\% with ME).
We had predicted either no effect or a gain of 2--5 points.
Since pairs share images, the CIs may be too narrow.
The first run stopped after the base model and ME$_1$ because the other versions' files were missing on the server; it was resumed with the same pairs before any analysis.
\subsection{New test sets}
\label{sec:suppl:newsets}
SPAQ-new are the SPAQ images at positions 5,000--6,999 of the fixed permutation whose first 5,000 positions served the pool-swap ablation; FLIVE-test is the whole official test split of FLIVE (1,827 images; we had planned 2,000 before finding that the split is smaller). Both are resized to a long side of 512 pixels. 9B was run on them after the other three models' results had been entered in the paper.

\begin{table}[h]
  \centering
  \footnotesize
  \setlength{\tabcolsep}{3pt}
  \caption{\textbf{Test sets never used for evaluation before} (SRCC, 12 held-out prompts; ranges over five versions; the KonIQ-fitted edits without refitting). Crit.: versions meeting the transfer criterion.}
  \label{tab:suppl:newsets}
  \begin{tabular}{llcccc}
    \toprule
    Data & Model & Base & ME & NME & Crit. \\
    \midrule
    \multirow{4}{*}{SPAQ} & 0.8B & 0.833 & 0.863--0.864 & 0.863--0.864 & 5/5 \\
     & 4B & 0.882 & 0.881--0.882 & 0.884--0.885 & 0/5 \\
     & Gemma & 0.781 & 0.850--0.852 & 0.849--0.851 & 5/5 \\
     & 9B & 0.866 & 0.865--0.867 & 0.864--0.866 & 0/5 \\
    \midrule
    \multirow{4}{*}{FLIVE} & 0.8B & 0.428 & 0.472--0.477 & 0.470--0.471 & 5/5 \\
     & 4B & 0.394 & 0.480--0.483 & 0.480--0.483 & 5/5 \\
     & Gemma & 0.321 & 0.450--0.461 & 0.450--0.457 & 5/5 \\
     & 9B & 0.387 & 0.479--0.484 & 0.479--0.483 & 5/5 \\
    \bottomrule
  \end{tabular}
\end{table}
\subsection{Qwen3.5-27B}
We used the official FP8 release, since the bfloat16 weights do not fit in 48\,GB; the last-layer MLP is converted to bfloat16 before editing, and, because the fused FP8 kernel needs a newer PyTorch than our server's, every FP8 weight is dequantized to bfloat16 before its matrix product.
The recipe is unchanged ($d_{\text{eff}}$ against the mean gradient: cosine 0.9999; pool selection SRCC 0.85--0.87); cached scores agree with real forward passes (Pearson 0.992--0.9996 over the 12 held-out prompts).
Under the 12 held-out prompts, SRCC moves from 0.782 to 0.853--0.857 on KonIQ (CIs above zero), 0.802 to 0.809--0.815 on LIVE Challenge and 0.806 to 0.829--0.834 on BID (CIs include or touch zero), and 0.723 to 0.611--0.615 on KADID (CIs below zero).
\subsection{Training-free alternatives}
\paragraph{Q-Debias.}
Q-Debias~\cite{chen2026qdebias} asks about the query image next to four synthetically distorted references and needs eight two-image forward passes per image.
We compare it with the base model and ME under its single-image question ``Rate the quality of the image. Good or poor?'' (\cref{tab:suppl:qd}).
It helps only on 0.8B KADID ($+$0.082, CI 0.048--0.118), where its references match the kind of distortion, and lowers KonIQ on both models; with it, ME scores lower than ME alone on all eight model--dataset pairs (the CI includes zero only on 0.8B BID).
Under this single question, ME lowers 4B's KADID SRCC from 0.698 to 0.498--0.523, much more than the 0.058 averaged over the 12 held-out prompts; per held-out prompt, the change on 4B KADID ranges from $-$0.206 to $+$0.064 (\cref{tab:suppl:full}). The KADID loss of the KonIQ-fitted edit thus depends strongly on the prompt.
\begin{table}[h]
  \centering
  \footnotesize
  \setlength{\tabcolsep}{2pt}
  \caption{\textbf{Q-Debias} (SRCC under the single-image question; ME: range over five versions).}
  \label{tab:suppl:qd}
  \begin{tabular}{llcccc}
    \toprule
    Model & Method & KonIQ & KADID & LIVE-C & BID \\
    \midrule
    \multirow{3}{*}{0.8B} & Base & 0.589 & 0.466 & 0.617 & 0.642 \\
     & Q-Debias & 0.534 & 0.549 & 0.535 & 0.650 \\
     & ME & 0.756--0.763 & 0.554--0.562 & 0.765--0.768 & 0.746--0.753 \\
    \midrule
    \multirow{3}{*}{4B} & Base & 0.792 & 0.698 & 0.788 & 0.803 \\
     & Q-Debias & 0.646 & 0.679 & 0.780 & 0.788 \\
     & ME & 0.818--0.821 & 0.498--0.523 & 0.789--0.792 & 0.837--0.840 \\
    \bottomrule
  \end{tabular}
\end{table}
\paragraph{Steering and few-shot prompting.}
\emph{Steering} adds $\alpha v$ at every position of the residual stream of layer $L$, where $v$ is the difference between the mean last-token states of the top and bottom MOS quartiles of the pool under three fit prompts, scaled to the layer's mean RMS; $(L,\alpha)$ was chosen from $\{6,12,18,23\}\times\{\pm1,\pm2,\pm4\}$ on 400 pool images ($L=18$, $\alpha=-4$).
Under the 12 held-out prompts it changes SRCC by $+$0.001 to $+$0.008 (KonIQ 0.649, base 0.642, ME 0.800), so a constant push does not reproduce the edit.
\emph{Few-shot prompting} shows $k$ pool images before the query, each followed by ``The quality of the image is good.'' (top 10\% MOS) or ``\dots poor.'' (bottom 10\%), alternating; under three held-out prompts, $k=2$ lowers SRCC by 0.07--0.20 and $k=4$ by 0.32--0.42 on the four datasets (all CIs below zero), whereas ME gains 0.04--0.17 under the same prompts.
On 4B, examples hurt more: SRCC falls from 0.72--0.81 to between $-$0.11 and 0.18 for both $k$ (all CIs below zero); we had predicted a gain of 0.02--0.05.
\subsection{Randomly drawn examples}
\label{sec:suppl:examples}
\begin{figure*}[t]
  \centering
  \includegraphics[width=\linewidth]{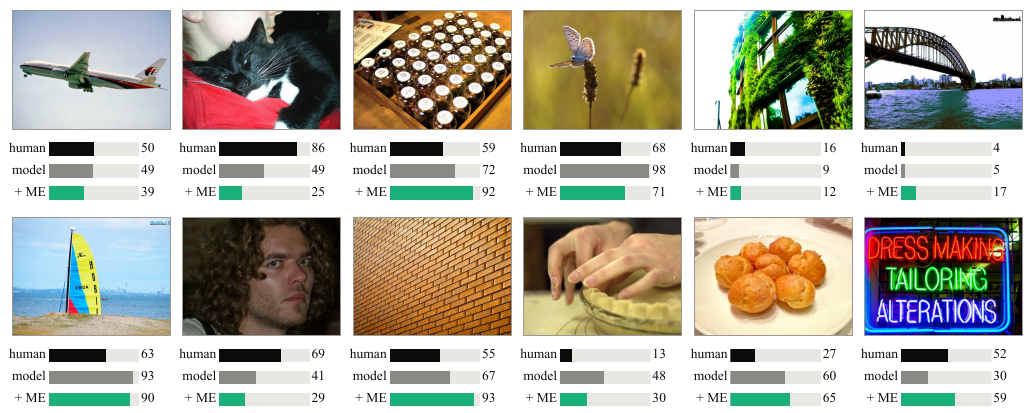}
  \caption{\textbf{Twelve KonIQ test images drawn at random} (Qwen3.5-0.8B). Bars: percentile among the 2,015 test images by MOS (human), by the base model's mean good$-$poor score over the 12 held-out prompts (model) and by ME (mean of five versions).}
  \label{fig:suppl:examples}
\end{figure*}

\Cref{fig:examples} shows images that the base answer misplaces by at least 40 percentiles.
\Cref{fig:suppl:examples} instead shows 12 KonIQ test images drawn uniformly at random, without any selection.
For these 12, the base model is within 15 percentiles of the human rank for 5 and ME for 5, and ME is closer than the base for 5; over all 2,015 test images, the base model is within 15 percentiles for 1,083 and ME for 1,390.
ME moves some images the wrong way (\eg, the second image: human 86, base 49, ME 25).

\section{Other Kinds of Images}
\label{sec:suppl:images}

\subsection{AI-generated images (AGIQA-3K)}
\label{sec:suppl:agiqa}
AGIQA-3K is split by text prompt into a pool (1,988 images) and a test set (994 images); ME and MTC are refit on the pool with the unchanged recipe, and CIs resample text prompts and held-out prompts.
\Cref{tab:suppl:agiqa} adds intervals to the AGIQA-3K column of \cref{tab:main}: all 20 versions are above the base model with the CI above zero, and the gain holds under 11 or 12 of the 12 held-out prompts.
The KonIQ-fitted edit does not transfer to these images ($+$0.001 to $+$0.005, not significant).

\begin{table*}[t]
  \centering
  \footnotesize
  \setlength{\tabcolsep}{4pt}
  \caption{\textbf{AGIQA-3K test set, edits refit on the AGIQA-3K pool} (994 images, 12 held-out prompts, five versions). CI: smallest lower and largest upper bound of the two-way 95\% CI of ME$-$Base over versions; Prompts: held-out prompts on which ME is above the base; MTC low: smallest lower CI bound of MTC$-$Base; ME$>$MTC: versions in which ME is significantly higher (MTC is never significantly higher).}
  \label{tab:suppl:agiqa}
  \begin{tabular}{lccccccccc}
    \toprule
    Model & Base & ME & ME$-$Base & CI & Prompts & Per-prompt gain & MTC & MTC low & ME$>$MTC \\
    \midrule
    0.8B & 0.687 & 0.825 & $+$0.137 & [0.097, 0.178] & 12 & $+$0.069 to $+$0.265 & 0.800 & 0.073 & 5/5 \\
    4B & 0.750 & 0.848 & $+$0.098 & [0.057, 0.139] & 11 & $-$0.011 to $+$0.225 & 0.836 & 0.049 & 5/5 \\
    Gemma & 0.623 & 0.818 & $+$0.195 & [0.123, 0.275] & 12 & $+$0.063 to $+$0.472 & 0.804 & 0.106 & 4/5 \\
    9B & 0.718 & 0.842 & $+$0.124 & [0.064, 0.181] & 11 & $-$0.057 to $+$0.276 & 0.837 & 0.072 & 0/5 \\
    \bottomrule
  \end{tabular}
\end{table*}

\subsection{Computer-graphics images (R4-CGQA)}
\label{sec:suppl:cg}
We use the ratings of R4-CGQA~\cite{li2026r4cgqa}, not its retrieval.
Scores are integers 0--5; the two images scored 0 are merged into 1, giving five levels.
We keep the authors' split: from the 3,190 library images we draw a pool of 2,000 (seed fixed before any model was run), the 90 fast-test images serve as validation, and the 220 test images are scored once.
Images are resized to a long side of 512. On these near-chance scores, 17 of the 528 per-prompt logistic fits for PLCC (mostly of base models) collapse to a two-valued step; we treat them as failed fits and use the raw scores, as for any failed fit. No other dataset in \cref{tab:main} has such fits.
The recipe, prompts and statistics are those of \cref{sec:exp:agiqa}; some held-out prompts mention ``photo'' and are used unchanged.
Gemma's test scoring was started locally and finished on a server after 5 of 11 versions; the merged scores keep the local values unchanged.

\begin{table*}[t]
  \centering
  \footnotesize
  \setlength{\tabcolsep}{6pt}
  \caption{\textbf{R4-CGQA test set} (220 images, 12 held-out prompts; ranges over 5 versions). KonIQ ME: the edit fitted on KonIQ, unchanged; refit: the same recipe fitted on 2,000 graphics images. CI: two-way bootstrap over images and prompts. Val: 90 validation images.}
  \label{tab:suppl:cg}
  \begin{tabular}{lccccc}
    \toprule
    & Base & KonIQ ME & \multicolumn{3}{c}{Refit on graphics} \\
    \cmidrule(lr){4-6}
    Model & SRCC / PLCC & $\Delta$ SRCC & SRCC / PLCC & $\Delta$ (CI low) & Val \\
    \midrule
    0.8B  & 0.109 / 0.172 & $-$0.007 to $+$0.006 & 0.36 / 0.38--0.39 & $+$0.25 (0.11--0.12) & 0.06\,$\to$\,0.41--0.42 \\
    4B    & 0.184 / 0.243 & $-$0.078 to $-$0.070 & 0.41--0.42 / 0.47--0.48 & $+$0.23 (0.09--0.10) & 0.17\,$\to$\,0.51 \\
    Gemma & 0.153 / 0.180 & $+$0.028 to $+$0.044 & 0.35--0.39 / 0.40--0.43 & $+$0.20--0.24 (0.08--0.11) & 0.13\,$\to$\,0.41--0.44 \\
    9B    & 0.148 / 0.278 & $+$0.012 to $+$0.020 & 0.38 / 0.46--0.47 & $+$0.23--0.24 (0.08) & 0.07\,$\to$\,0.47--0.48 \\
    \bottomrule
  \end{tabular}
\end{table*}

The KonIQ-fitted edit does not transfer: all 20 of its confidence intervals include zero, and on 4B the point estimate is negative.
Refit on graphics, all 20 versions improve with a CI above zero and all 12 prompts improving, but the CIs are wide (upper bounds 0.32--0.39) and the best SRCC is 0.42, so the refit edit makes the models better, not good, judges of graphics.

\paragraph{Reference: models trained on photographs.}
The public Q-ReAlign models were trained with the Q-Align recipe on KonIQ, SPAQ, KADID, AVA and LSVQ and never saw graphics.
We score the same 220 test images with their own prompt and level-token score, at our long side of 512 and at 1,024 (resized from the authors' originals, closer to their native resolution), and under our 12 held-out prompts (\cref{tab:suppl:cgref}).
At 1,024 they reach 0.23--0.30, 0.08--0.13 below the refit edit on the base model of the same size; for Mini and Pro the differences are not significant (95\% CI upper bounds 0.010--0.037), whereas Lite is significantly below the refit edit in four of the five paired versions (upper bounds $-$0.005 to $+$0.004).
The comparison is not matched: the refit edit uses 2,000 graphics labels and Q-ReAlign none, so we report it only as a reference for what photograph training gives on graphics.

\begin{table}[h]
  \centering
  \footnotesize
  \setlength{\tabcolsep}{2pt}
  \caption{\textbf{Q-ReAlign on the R4-CGQA test set.} Own: Q-Align prompt and level-token score (SRCC / PLCC); held-out: mean SRCC under our 12 prompts; refit: our edit refit on graphics, same base size (12 held-out prompts, mean of 5 versions).}
  \label{tab:suppl:cgref}
  \begin{tabular}{lcccc}
    \toprule
    & Own @512 & Own @1,024 & Held-out @1,024 & Refit \\
    \midrule
    Mini (0.8B) & 0.126 / 0.106 & 0.232 / 0.231 & 0.079 & 0.360 \\
    Lite (4B)   & 0.170 / 0.127 & 0.293 / 0.321 & 0.102 & 0.416 \\
    Pro (9B)    & 0.270 / 0.289 & 0.299 / 0.329 & 0.013 & 0.382 \\
    \bottomrule
  \end{tabular}
\end{table}

\subsection{Choosing among generated images}
\label{sec:suppl:bestofn}
The AGIQA-3K test set has 100 text prompts with 7--10 generated images each.
For each prompt we select the image with the highest good$-$poor score (mean over the 12 held-out prompts) and report, averaged over the 100 prompts, how often it is the image with the highest MOS (top-1; chance 0.10) and its MOS percentile within the prompt (chance 0.5), with ME and MTC refit on the AGIQA pool (\cref{tab:suppl:bestofn}).
For every model and every version, ME $-$ base has a 95\% CI above zero on both metrics and on pairwise ranking within prompts (bootstrap over text prompts); MTC improves as well, and no MTC version is significantly above ME.
MOS is not a direct choice between generations, and images of one prompt come from different generators.

\begin{table}[h]
  \centering
  \footnotesize
  \caption{\textbf{Best-of-$N$ selection on AGIQA-3K} (100 test prompts; ME and MTC: mean of five versions).}
  \label{tab:suppl:bestofn}
  \begin{tabular}{lcccccc}
    \toprule
    & \multicolumn{3}{c}{Top-1} & \multicolumn{3}{c}{MOS percentile} \\
    \cmidrule(lr){2-4} \cmidrule(lr){5-7}
    Model & Base & MTC & ME & Base & MTC & ME \\
    \midrule
    0.8B  & 0.24 & 0.43 & 0.51 & 0.77 & 0.88 & 0.90 \\
    4B    & 0.41 & 0.54 & 0.60 & 0.83 & 0.90 & 0.91 \\
    Gemma & 0.18 & 0.51 & 0.53 & 0.73 & 0.89 & 0.90 \\
    9B    & 0.31 & 0.51 & 0.57 & 0.78 & 0.90 & 0.91 \\
    \bottomrule
  \end{tabular}
\end{table}

\paragraph{Probes as judges.}
To ask whether an external read-out would choose better, we fitted ridge probes on the base model's states for the AGIQA pool images under the first held-out prompt: one on the last-token last-layer state, and, for the Qwen models, one on the patch mean of a vision-encoder block, with the penalty and the block chosen by 5-fold cross-validation grouped by text prompt.
Each probe scored the test images once (\cref{tab:suppl:probejudge}).
Of the 105 paired CIs of probe $-$ ME (7 probes, 5 versions, 3 metrics), 102 include zero; the other three favor Gemma's last-layer probe on pairwise ranking by at most 0.023.
We had predicted the vision probes to pick better than ME; they did not.

\begin{table}[h]
  \centering
  \footnotesize
  \setlength{\tabcolsep}{3pt}
  \caption{\textbf{Probes vs.\ ME as judges on AGIQA-3K} (top-1 / MOS percentile; ME: range over five versions; Gemma: no vision probe).}
  \label{tab:suppl:probejudge}
  \begin{tabular}{lccc}
    \toprule
    Model & ME & Last-layer probe & Vision probe \\
    \midrule
    0.8B  & 0.50--0.52 / 0.90 & 0.55 / 0.90 & 0.52 / 0.89 \\
    4B    & 0.59--0.60 / 0.91 & 0.60 / 0.91 & 0.57 / 0.92 \\
    Gemma & 0.52--0.53 / 0.89--0.90 & 0.53 / 0.90 & -- \\
    9B    & 0.57 / 0.91 & 0.56 / 0.89 & 0.54 / 0.91 \\
    \bottomrule
  \end{tabular}
\end{table}
\section{Training the Whole Model}
\label{sec:suppl:train}
\begin{table}[t]
  \centering
  \footnotesize
  \setlength{\tabcolsep}{2.4pt}
    \caption{\textbf{ME vs.\ training the whole model.}
    Own: each trained model's Q-Align prompt and level-token score (Q-ReAlign at 512\,/\,1,024 resolution); held-out: good$-$poor under our 12 prompts; LIVE-C: own score for trained models.
    Q-Align (2k or 7k KonIQ labels; LoRA on 4B): mean of three seeds, held-out for seed 1 (0.8B, 2k: own 0.890, 0.895, 0.895 over seeds, so the equal own and held-out values are a coincidence of rounding).
    KL: Q-Align LoRA (2k) with a per-token KL penalty, weight $\beta_{\text{KL}}$, to the frozen base model's greedy answers to four generic questions on pool images; three seeds (held-out: mean of three), $^\S$one seed.
    LLVQA: change in LLVisionQA accuracy (points).}
  \label{tab:train}
  \resizebox{\linewidth}{!}{%
  \begin{tabular}{llccccc}
    \toprule
    & & \multicolumn{2}{c}{KonIQ} & & LLVQA & Facts \\
    \cmidrule(lr){3-4}
    Model & Method & own & held-out & LIVE-C & $\Delta$ & /40 \\
    \midrule
    \multirow{5}{*}{0.8B} & Base & 0.506 & 0.642 & 0.694 & -- & 38 \\
      & +ME & -- & 0.800 & 0.781 & $+$0.1 & 38 \\
      & Q-Align, 2k & 0.893 & 0.893 & 0.817 & $-$25.6 & 0 \\
      & Q-Align, 7k & 0.919 & 0.834 & 0.832 & $-$25.3 & 0 \\
      & Q-ReAlign & 0.769\,/\,0.910 & 0.530 & 0.793 & $-$22.5 & 0 \\
    \midrule
    \multirow{5}{*}{4B} & Base & 0.740 & 0.742 & 0.779 & -- & 40 \\
      & +ME & -- & 0.842 & 0.810 & $-$0.1 & 40 \\
      & Q-Align, 2k & 0.903 & 0.731 & 0.867 & $-$10.8 & 39--40 \\
      & Q-Align, 7k & 0.923 & 0.772 & 0.878 & $-$23.5 & 37--39 \\
      & 2k + KL, $\beta_{\text{KL}}{=}1$ & 0.903 & 0.849 & 0.861 & $+$3.1 & 39--40 \\
      & 2k + KL, $\beta_{\text{KL}}{=}0.1^\S$ & 0.904 & 0.798 & 0.870 & $+$1.7 & 40 \\
      & Q-ReAlign & 0.775\,/\,0.943 & 0.606 & 0.868 & $-$35.9 & 0 \\
    \midrule
    \multirow{3}{*}{9B} & Base & -- & 0.709 & 0.794 & -- & 40 \\
      & +ME & -- & 0.834 & 0.816 & $+$0.1 & 40 \\
      & Q-ReAlign & 0.809\,/\,0.950 & 0.574 & 0.880 & $-$43.8 & 0 \\
    \bottomrule
  \end{tabular}}
\end{table}
\begin{figure}[t]
  \centering
  \includegraphics[width=\linewidth]{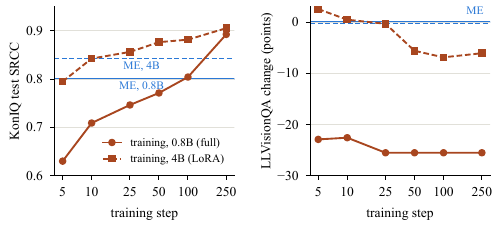}
  \caption{\textbf{What training gains and loses along the way.}
    One Q-Align-style training run on the 2,000 pool labels (seed 1), evaluated at steps 5--250: KonIQ test SRCC under the model's own scoring (left) and the change in LLVisionQA accuracy (right).
    Blue lines: ME on the same base model (solid 0.8B, dashed 4B).
    Full fine-tuning of 0.8B loses LLVisionQA accuracy within five steps, long before KonIQ SRCC reaches ME's level; LoRA on 4B matches ME on both at steps 10 and 25 and loses accuracy from step 50.}
  \label{fig:curve}
\end{figure}

\paragraph{Resolution.}
Resolution does not explain Q-ReAlign's lead on KonIQ: at 1,024$\times$768, where ME raises KonIQ SRCC from 0.707 to 0.835 (0.8B) and from 0.778 to 0.862 (4B), Q-ReAlign still leads by 0.075 and 0.082.

\paragraph{Choosing the stopping step without the test set.}
\Cref{fig:curve} follows one run step by step. For 4B LoRA (seed 1) we scored the six saved checkpoints on the 1,000 KonIQ validation images with the model's own score (\cref{tab:suppl:lora}).
Validation SRCC rises at every checkpoint, so stopping at the best validation SRCC selects the last step (250), which loses 6.1 LLVisionQA points.
Stopping at the first checkpoint whose validation SRCC reaches ME's (0.841, 12 held-out prompts, mean of five versions) selects step 10, which matches ME on both KonIQ test SRCC and LLVisionQA.
This rule needs ME's validation score as its target.

\begin{table}[h]
  \centering
  \footnotesize
  \setlength{\tabcolsep}{4pt}
  \caption{\textbf{4B LoRA along training} (seed 1). Validation and test: KonIQ SRCC under the model's own score; LLVQA: change in LLVisionQA accuracy (points). ME: 0.841 validation, 0.842 test.}
  \label{tab:suppl:lora}
  \begin{tabular}{lcccccc}
    \toprule
    Step & 5 & 10 & 25 & 50 & 100 & 250 \\
    \midrule
    Validation & 0.804 & 0.846 & 0.854 & 0.878 & 0.884 & 0.903 \\
    Test & 0.795 & 0.843 & 0.856 & 0.877 & 0.882 & 0.906 \\
    LLVQA $\Delta$ & $+$2.5 & $+$0.5 & $-$0.3 & $-$5.6 & $-$6.9 & $-$6.1 \\
    \bottomrule
  \end{tabular}
\end{table}
\section{Side Effects and Constraints}
\label{sec:suppl:effects}

\subsection{Side-effect details}
\label{sec:suppl:side}
Besides \cref{tab:side}, we ran a small set of text probes and image descriptions on three ME versions per model (greedy decoding).
\emph{Facts}: 40 short factual questions (\eg, capitals, arithmetic).
\emph{Neutral}: 10 neutral sentence starts (\eg, ``The sun rises in the''), checking whether the most likely next token changes.
\emph{Valence}: 10 pairs of positive and negative reviews ending in a verdict to complete (\eg, ``The food was cold and the staff were rude. Overall, the meal was''); we compare the log-probabilities of ``good'' and ``poor'' and count sentences where the larger one matches the sentence's valence.
\emph{Descriptions}: 400 open image descriptions, checked for degenerate repetition and compared with the base model's.
\begin{table}[h]
  \centering
  \footnotesize
  \setlength{\tabcolsep}{3pt}
  \caption{\textbf{Text probes and descriptions} (base $\to$ three ME versions). Neutral: sentences whose top next token is unchanged. Val.\ match: sentences whose ``good''/``poor'' preference matches their valence (of 20; not the valence gap of \cref{tab:side}). Degenerate: descriptions that repeat one word to the length limit.}
  \label{tab:suppl:side}
  \begin{tabular}{lcccc}
    \toprule
    Model & Facts /40 & Neutral /10 & Val.\ match /20 & Degenerate \\
    \midrule
    0.8B  & 38\,$\to$\,38 & 9--10 & 16\,$\to$\,10 & 0\,$\to$\,0.8--1.3\% \\
    4B    & 40\,$\to$\,40 & 10 & 18\,$\to$\,11 & 0\,$\to$\,0 \\
    Gemma & 40\,$\to$\,40 & 4--5 & 17\,$\to$\,12--13 & 0\,$\to$\,5.5--13\% \\
    9B    & 40\,$\to$\,40 & 9--10 & 17\,$\to$\,13--14 & 0\,$\to$\,0.25\% \\
    \bottomrule
  \end{tabular}
\end{table}
On 0.8B, the base model's mean ``good''$-$``poor'' log-probability is $+$6.4 after positive and $-$0.1 after negative sentences; after ME both are about $+$3, so the edit removes most of this difference (\cref{tab:side}), while the top next token of these sentences barely changes (``a'' and ``very'' remain the most frequent).
Only 17--22\% of 0.8B descriptions are identical to the base model's, and quality words appear in 11.5--13.5\% of them instead of 8\%.
Gemma's degenerate descriptions repeat ``poor'' (\eg, 51 of 400 for one version).
On LLVisionQA (1,495 multiple-choice questions, dev split), ME's answers agree with the base model's on 98.7--99.5\% of questions for all four models, and accuracy changes by at most about one point on each of its four question concerns.

\paragraph{General benchmarks.}
We scored, without generation, perplexity on the first 64 non-overlapping 1,024-token segments of the WikiText-2 test set (plain text), SST-2 validation accuracy (872 sentences; ``Is the sentiment of this sentence positive or negative?'', prefix ``The sentiment is'', `` positive'' vs.\ `` negative''), and MMBench English dev accuracy (4,329 single-choice questions, highest option-letter logit), with paired bootstrap CIs over segments or questions (\cref{tab:suppl:bench}).

\begin{table}[h]
  \centering
  \footnotesize
  \setlength{\tabcolsep}{3pt}
  \caption{\textbf{General benchmarks} (versions 1/2/3; Gemma: ranges). On 0.8B, all perplexity increases and SST-2 drops of ME have CIs excluding zero; MMBench CIs include or touch zero. On 4B and 9B, all perplexity increases have CIs excluding zero. For Gemma, plain-text perplexity was not stable to measure (base above 190 even with a BOS token) and is omitted.}
  \label{tab:suppl:bench}
  \begin{tabular}{lccc}
    \toprule
    & Perplexity & SST-2 (\%) & MMBench (\%) \\
    \midrule
    \multicolumn{4}{l}{\emph{Qwen3.5-0.8B}} \\
    Base & 19.82 & 86.7 & 75.2 \\
    ME   & 20.57 / 20.23 / 20.11 & 83.3 / 83.0 / 84.2 & 75.6 / 75.5 / 75.3 \\
    MTC  & 20.99 / 20.56 / 20.15 & 84.3 / 85.7 / 86.2 & 75.3 / 75.1 / 75.3 \\
    \multicolumn{4}{l}{\emph{Qwen3.5-4B}} \\
    Base & 10.80 & 93.8 & 86.2 \\
    ME   & 10.87 / 10.93 / 10.86 & 93.9 / 93.9 / 93.8 & 86.3 / 86.2 / 86.3 \\
    MTC  & 10.82 / 10.86 / 10.82 & 93.8 / 93.7 / 93.8 & 86.3 / 86.2 / 86.4 \\
    \multicolumn{4}{l}{\emph{Qwen3.5-9B}} \\
    Base & 9.29 & 94.8 & 90.5 \\
    ME   & 9.38 / 9.35 / 9.37 & 95.1 / 95.1 / 95.2 & 90.5 / 90.5 / 90.5 \\
    MTC  & 9.37 / 9.43 / 9.33 & 95.0 / 94.8 / 94.8 & 90.6 / 90.5 / 90.5 \\
    \multicolumn{4}{l}{\emph{Gemma 4 E2B}} \\
    Base & -- & 87.2 & 68.9 \\
    ME   & -- & 87.3 & 68.8--68.9 \\
    MTC  & -- & 87.3--87.7 & 68.7--69.0 \\
    \bottomrule
  \end{tabular}
\end{table}

We had predicted a perplexity change below 2\% (0.8B ME$_1$: 3.8\%; 4B and 9B: 0.5--1.3\%) and an SST-2 drop that could exceed 2 points for 0.8B's ME (it did; on the other models, ME changes SST-2 and MMBench by at most 0.4 points).
\paragraph{Why completed answers become more positive.}
On 0.8B under ``Rate the quality of the image.'' / ``The quality of the image is'', every base answer begins with `` **'' before the quality word.
We measured, for the base model and ME$_1$--ME$_3$, the good$-$poor logit gap at the end of each fit prompt on the KonIQ pool, and the next-token log-probabilities of six quality words after `` **'' on the pool and on the worst test quartile (after ``**'', ``poor'', ``decent'' and ``excellent'' begin with the word pieces ``po'', ``de'' and ``ex'').
Averaged over the 12 fit prompts, ME moves the gap by $-$0.10 to $+$0.20 logits, but under this prompt by $+$1.1 to $+$1.5 (CIs above zero), and across single fit prompts by $-$7.6 to $+$9.8.
At the position of the quality word, the gap moves by a similar $+$1.1 to $+$1.6 on the pool, and ``good'' becomes the most likely of the six words for 36--42\% of the worst quartile instead of 10\%, matching the 9\% to 37\% of the generated answers.
In that quartile, ``good'' gains on ``low'' and ``decent'' ($+$0.43 to $+$0.72) but also on ``poor'' ($+$0.34 to $+$0.80), so the plan's second explanation, that only the contrast with competing words moves, did not hold; we had predicted it would.

\paragraph{Side effects of the refit edits.}
With the protocol of \cref{tab:side}, the edits refit on AGIQA-3K and on R4-CGQA change LLVisionQA accuracy by $-$0.2 to $+$0.3 points on all four models (all 40 CIs include zero).
Factual answers are unchanged except on Gemma, where two of the five graphics edits answer 38 of 40 instead of 40, and on 0.8B, where they rise from 38 to 39--40.
The graphics edits shrink the ``good''$-$``poor'' valence gap like the KonIQ edit (\eg, 9B: 8.2 to 2.2--2.4; Gemma: 5.4 to 0.1--1.2; base values re-measured in the same run), whereas the AGIQA edit on 4B enlarges it (7.4 to 8.9--9.3).
\subsection{Text-side constraint: projection and penalty}
\label{sec:suppl:textpen}
Instead of projecting out the last-token states of 130 sentences (NME), one can add $\mu\sum_t(a_k(x_t)\,c^\top x_t)^2$ over every token of 64 WikiText-2 validation segments and the same sentences, with $\mu$ chosen on the pool and on held-out text.
On 0.8B this trades one property for the other (version 1, KonIQ SRCC / valence gap on the ten evaluation pairs): with $\mu = 10^{k}$ times the ratio of the image and text Gram traces, 0.801\,/\,0.18 without penalty, 0.799\,/\,0.91 at $k=0$, 0.787\,/\,3.78 at $k=1$, 0.742\,/\,6.58 at $k=2$ and 0.683\,/\,7.22 at $k=3$, whereas NME gives 0.800\,/\,6.2--6.3.
The rule fixed before testing (smallest $\mu$ keeping 90\% of the held-out valence with a pool score within 0.01 of the unpenalized one) found no such $\mu$ and fell back to its second rule, which selected $k=1$ for four versions.
\paragraph{A generic-text constraint (NME-G).}
\label{sec:suppl:nmeg}
Because NME's 130 sentences share the format of the ten evaluation reviews, we fixed before computing a second evaluation and a second constraint.
The evaluation uses all 872 SST-2 validation sentences under the instruction ``Continue the text.'', each followed by ``In one word, I found it'', and takes the log-probability of `` good'' minus `` poor'' at the next token; the SST-2 gap is its mean over positive minus negative sentences.
The constraint takes the last-MLP inputs at every position from 8 on of 40 windows of 128 tokens from the first half of the WikiText-103 validation articles (4,800 states), and projects out the top-$r$ eigenvectors of their uncentered second moment together with the pool mean, as in AlphaEdit~\cite{fang2025alphaedit}.
Among $r \in \{16, 32, 64, 128, 256, 512\}$ whose pool selection SRCC is within 0.01 of ME's, we took the one with the lowest next-token KL to the base model on 16 windows of 256 tokens from the second half of the articles: $r = 128$ on 0.8B and $r = 512$ on 4B and 9B.
The constraint would count as holding on a model if, averaged over the five versions, KonIQ SRCC stayed within 0.01 of ME, SST-2 accuracy within one point of the base model, the SST-2 gap at 90\% of the base model's or more, and perplexity within 1\%.
\Cref{tab:suppl:nmeg} gives the results; 0.8B fails the last three conditions, 4B and 9B the third.
Over $r$ (version 1), the SST-2 gap on 0.8B is $-$0.13, $-$0.13, $-$0.09, 0.29, 0.90 and 0.36, on 4B $-$1.64, $-$1.57, $-$1.44, $-$1.52, $-$1.44 and $-$0.97, and on 9B $-$0.39, $-$0.32, $-$0.23, $-$0.14, $-$0.06 and $-$0.46, with KonIQ SRCC 0.788--0.804, 0.839--0.843 and 0.826--0.830 and SST-2 accuracy 80.3--81.3\%, 93.7--93.9\% and 95.0--95.2\%.

\begin{table}[h]
  \centering
  \footnotesize
  \setlength{\tabcolsep}{2.5pt}
  \caption{\textbf{Text-side constraints} (means of five versions; base, ME and NME re-measured in the same run as NME-G; SST-2 gap with 95\% CI over sentences; Val.: ten-review gap; NME's KonIQ is in \cref{tab:side}).}
  \label{tab:suppl:nmeg}
  \begin{tabular}{llccccc}
    \toprule
    & & KonIQ & SST-2 (\%) & SST-2 gap & PPL & Val. \\
    \midrule
    \multirow{4}{*}{0.8B} & base & 0.642 & 86.4 & 1.75 [1.59, 1.93] & 19.81 & 6.50 \\
     & ME & 0.800 & 83.2 & $-$0.62 [$-$0.98, $-$0.26] & 20.24 & 0.21 \\
     & NME & -- & 80.4 & 1.52 [1.22, 1.83] & 20.17 & 6.23 \\
     & NME-G & 0.803 & 81.0 & 0.29 [$-$0.02, 0.60] & 20.08 & 1.44 \\
    \midrule
    \multirow{4}{*}{4B} & base & 0.742 & 93.8 & 2.25 [2.09, 2.42] & 10.79 & 7.39 \\
     & ME & 0.842 & 93.8 & $-$1.97 [$-$2.31, $-$1.62] & 10.87 & 3.47 \\
     & NME & -- & 94.1 & 0.53 [0.14, 0.92] & 10.88 & 8.32 \\
     & NME-G & 0.840 & 93.7 & $-$1.12 [$-$1.48, $-$0.76] & 10.85 & 5.62 \\
    \midrule
    \multirow{4}{*}{9B} & base & 0.709 & 94.8 & 1.58 [1.44, 1.72] & 9.29 & 8.25 \\
     & ME & 0.834 & 95.1 & $-$0.57 [$-$0.90, $-$0.21] & 9.36 & 5.59 \\
     & NME & -- & 94.5 & 0.34 [0.00, 0.70] & 9.37 & 8.66 \\
     & NME-G & 0.832 & 95.0 & $-$0.53 [$-$0.87, $-$0.18] & 9.35 & 4.71 \\
    \bottomrule
  \end{tabular}
\end{table}
\paragraph{A norm constraint.}
A dense rank-one variant of ME whose update norm was capped at 5\% of the down-projection's (CME, 0.8B, one version, real forward passes) kept a valence gap of only 1.0 (ME: about 0.2) at a KonIQ SRCC of 0.768 (ME$_1$: 0.801).

\paragraph{An edit in the vision encoder.}
Agent mo2 (\cref{sec:suppl:agentruns}) changed one MLP neuron of vision block 9 of 0.8B (2,305 nonzero weight changes), so text-only inputs never pass through the edited weights.
In one run with the base model, its SST-2 accuracy (86.4\%) and SST-2 gap (1.75, CI [1.59, 1.93]) equal the base model's exactly (largest difference over the 872 sentences: 0.0); its valence gap is 6.5 and its LLVisionQA accuracy $+$0.7 points (\cref{tab:suppl:mo}).
It reaches 0.759 on KonIQ against ME's 0.801 under the same real forward passes, and does not improve KADID (0.545, base 0.548).

\input{sec/4_agents}

\subsection{Agent protocol and prompts}
\begin{figure}[h]
  \centering
  \includegraphics[width=\linewidth]{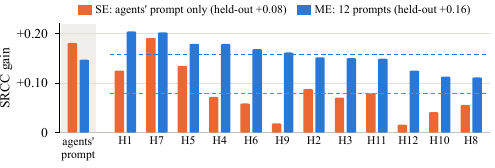}
  \caption{\textbf{One prompt or twelve.}
    SRCC gain of Qwen3.5-0.8B over the base model on the KonIQ test set, under the agents' prompt (shaded) and the 12 held-out prompts, sorted by ME's gain; dashed lines: held-out means.
    SE and ME share neurons and closed form and differ only in the fitting objective.}
  \label{fig:objective}
\end{figure}
\label{sec:suppl:agents}
\paragraph{Agents.}
All 19 editing agents in this paper (run1--run5, lf1--lf3, ll25, ll100, agiqa1--agiqa3, mo1--mo3 and qa1--qa3) were Claude Opus 5.5 (model id \texttt{claude-opus-5-5}) running as Claude Code subagents with default sampling settings, one session each, with no access to each other's work and no way to ask questions.
Each could run shell commands and write files in its own run directory; it was told not to use the internet or start further agents.
We did not repeat any agent run, so the variation between the five runs of \cref{sec:agents} is the only measure of run-to-run randomness; the runs differ also in budget (50, 50, 50, 25 and 10 calls).

\paragraph{Task brief.}
The five agents of \cref{sec:agents} received the same brief, which differs only in the run directory and budget.
It asks the agent to edit the weights of Qwen3.5-0.8B so that the score of \cref{eq:score} under ``Rate the quality of the image.'' (prefix ``The quality of the image is'', thinking disabled) ranks unseen images closer to MOS, to diagnose before editing, and to report a negative result honestly if no edit is worthwhile.
Its rules, in brief:
(1)~the model may be run only through the provided tool, at most 256 images per call, every call counted and logged;
(2)~an edit is a file of additive weight deltas, allowed only in language-model layers and vision blocks (not embeddings, final norm or merger);
(3)~no gradients, iterative optimizers or search that fits the edit to labels; closed-form statistics are allowed, and any direction computed from MOS must be declared ``label-fitted'' with the number of labels used;
(4)~read only the shared pool (2,000 KonIQ images with MOS), the weights and one's own directory, never test data, results or other runs;
(5)~hold out pool images to check generalization;
(6)~prefer edits that change few weights and note what else they may disturb.
The submission is one edit, notes (diagnosis, every attempt with its images and SRCC, calls used, doubts) and every script needed to rebuild the edit.
\Cref{fig:suppl:brief} reproduces the brief verbatim; the other runs below changed only the goal, data and rules named there.
\begin{figure*}[t]
  \centering
  \fbox{\parbox{0.97\textwidth}{\scriptsize\raggedright
    You are the editor in a neural-network weight-editing experiment. Work independently and finish on your own; nobody can answer questions.\par
    \par\smallskip
    \par\textbf{GOAL}\par
    Directly edit the weights of Qwen3.5-0.8B (a small vision-language model) so its image-quality score ranks images closer to human opinion scores (Spearman SRCC with MOS) on images you never see. Score = logit(`` good'') - logit(`` poor'') at the next-token position after this fixed prompt: an image, ``Rate the quality of the image.'', then the assistant prefix ``The quality of the image is'' (thinking disabled). A constant offset to this score cannot change SRCC, so the edit has to make it vary differently across images. Diagnose first and find out why the score ranks poorly, then design the edit from that diagnosis. If you find no worthwhile edit, say so honestly and submit the best you have or no edit; a truthful negative result is a useful outcome.\par
    \par\smallskip
    \par\textbf{PATHS}\par
    EXP=\textless{}experiment folder\textgreater{}\par
    Your run directory: \$EXP/editor/\textless{}run\textgreater{} (forward-pass budget: \textless{}B\textgreater{} calls)\par
    Python: \textless{}virtual environment\textgreater{} (always run with -I; it has torch, transformers, safetensors, numpy, scipy, sklearn, pandas)\par
    Tool: \$EXP/editor/shared/forward.py (read it; its --help lists the arguments)\par
    Labelled images: \$EXP/editor/shared/pool.csv (pool\_index, img\_name, mos; 2,000 images, in editor/shared/images)\par
    Weights, read-only: \$EXP/models/qwen35\_08b/model.safetensors-00001-of-00001.safetensors (tensors are bfloat16: open with safetensors framework=``pt'', then .float().numpy()). Parameter names are as in the file, e.g. model.language\_model.layers.15.mlp.down\_proj.weight with shape [1024, 3584].\par
    \par\smallskip
    \par\textbf{RULES}\par
    1. Run the model only through forward.py. Each call costs 1 of your budget, handles at most 256 images, and is logged. Do not load the model any other way, on CPU or GPU, and do not alter budget or ledger files.\par
    2. An edit is an .npz mapping parameter names to delta arrays of the full parameter shape, added to the weights. Only model.language\_model.layers.N.* and model.visual.blocks.N.* may change; the tool rejects anything else, including embed\_tokens, final norm and the merger. Do not try to get around this.\par
    3. No gradients, backpropagation, or iterative optimizers or search that fits the edit to labels (no SGD, no hill climbing over many candidates by SRCC). You may freely analyse weights and captured activations with Python and closed-form statistics (means, PCA/SVD, projections). You may evaluate a few hand-reasoned candidates with your budget, and you may use MOS labels for analysis. If any part of your final edit's direction was computed from MOS labels (e.g. a difference of means or closed-form regression), that is permitted but you MUST declare it plainly in your notes as ``label-fitted'', with how many labelled images it used.\par
    4. Touch only your own run directory for writing. Read only editor/shared, your own run directory and the model files. Do NOT read other editor runs, \$EXP/results, \$EXP/data, \$EXP/code, or \$EXP/EXPERIMENT\_DESIGN.md: those hold protected evaluation data. Do not use the internet or spawn subagents.\par
    5. To check generalization, keep some pool images out of whatever you use to build the edit and evaluate on those. Beware: small differences on 256 images are noisy, and repeatedly picking winners on the same images overstates results.\par
    6. Prefer edits that change as few weights as practical and that are less likely to damage the model's other abilities. The edit applies to all inputs, not only this prompt, so note what else it may disturb. You will be evaluated on side effects as well.\par
    7. The GPU is shared and forward.py waits for a lock, so a call may take a few minutes. Use a generous timeout (up to 600000 ms) and run calls one at a time. Keep total wall-clock time reasonable, and stop when more calls no longer teach you anything; you need not spend the full budget.\par
    \par\smallskip
    \par\textbf{SUBMISSION, required}\par
    - \$EXP/editor/\textless{}run\textgreater{}/submit/edit.npz: your single final edit (omit if you submit no edit)\par
    - \$EXP/editor/\textless{}run\textgreater{}/submit/notes.md: diagnosis; what you changed and the number of weights; the method for each direction, stating whether label-fitted and with how many labels; all attempts including failures, each with its images and SRCC; calls used; your honest assessment and doubts.\par
    - Every script needed to rebuild the edit, in your run directory.\par
    \par\smallskip
    Final reply: under 150 words stating what you submitted, the calls used, your held-out estimate, and whether it is label-fitted.\par
  }}
  \caption{\textbf{Task brief of the five agents in \cref{sec:agents}}, verbatim except that local paths are replaced by placeholders; \textless{}run\textgreater{} and \textless{}B\textgreater{} are run1--run5 and 50, 50, 50, 25, 10.}
  \label{fig:suppl:brief}
\end{figure*}

\paragraph{Tool.}
\texttt{forward.py} takes a run directory, a list of pool indices, an optional edit and the states to capture: hidden states of any language layer, outputs of any vision block, and inputs and outputs of named language modules.
It returns the score for every image and the requested arrays, which the agent may analyze offline without limit.
Each submission was scored once on the locked KonIQ test set by us, after the agent had finished.

\paragraph{Other agent runs.}
\label{sec:suppl:agentruns}
Three agents (lf1--lf3) received the same task without any labels, with 2,000 unlabeled photos and permission to create synthetic distortions; none raised SRCC by 0.03 on any dataset (\cref{sec:agents}).
Two received only 25 or 100 labels (ll25, ll100); their edits reached 0.712 and 0.717 KonIQ SRCC, below a ridge probe on the same 25 labels (0.767), and are not used elsewhere.
Three (agiqa1--agiqa3) received the brief of \cref{sec:agents} with the AGIQA-3K pool in place of KonIQ (\cref{sec:exp:agiqa}).
Three (mo1--mo3) worked under a constraint (below).
Three (qa1--qa3) worked on the 549 yes/no questions of LLVisionQA, where a probe on 0.8B's last-token states answers more often correctly than the model (75.2\% vs.\ 63.0\%); they raised accuracy on 275 locked questions from 63.6\% to 69.5--70.9\% by cancelling the model's preference for ``A'' and ``yes'', which a label-free calibration of the output that changes no weight matches (70.9\%; paired CIs include zero), so we do not report them further.

\paragraph{Search without an agent.}
Two scripted searches had the agents' information and budget: their prompt, the 2,000 labeled pool images, at most 50 calls of 256 images, and no gradients.
Both fitted the same edit everywhere: one MLP neuron of layer $L$ replaced by a ridge read-out written along $d_{\text{eff}}$, with ME's closed form for a single prompt, from the last-token MLP inputs of 1,280 pool images captured in five calls.
The sweep (S) fitted a neuron with a stable gate in each of the 24 layers and checked each edit with one call on 256 other pool images; random search (R) did the same for 45 random (layer, neuron) pairs.
Each submitted its best edit, scored once on the locked KonIQ test set.
S chose layer 15 after 29 calls (layers 14--19 and 23--24 all checked at 0.82--0.83, the last layer at 0.827) and reached 0.835 under the agents' prompt (base 0.652 in this run; 0.653 in the agents' audit on the same images, a difference we did not trace); R chose a neuron of layer 18 and reached 0.833.
Under the nine prompts no agent saw, their mean gains were $+$0.037 (CI $+$0.003 to $+$0.075) and $+$0.063 (CI $+$0.037 to $+$0.091), against $-$0.02 to $+$0.07 for the agents.
We had predicted that S would choose the last layer and that R would score below S; neither held.

\paragraph{Agents under a constraint.}
Three agents (mo1--mo3) received the 12 fit prompts, the labeled pool and a brief asking for better quality ranking under unseen prompts while keeping the ``good''/``poor'' contrast in plain text and other abilities; a good trade-off was to matter more than the best value on one.
Besides the image tool (50 calls), a text tool returned the ``good''/``poor'' logits and top-5 next tokens for up to 256 text-only items that the agent wrote itself; the evaluation sentences were hidden.
The criterion, fixed before testing, counted an agent as successful if (a) KonIQ SRCC rose by at least 0.10 with the paired CI above zero, (b) the valence gap stayed at least 3.25 (half the base value), and (c) facts stayed within two of the base model and LLVisionQA within one point; it held if at least two of three agents succeeded.
mo1 and mo2 met all three; mo3 missed (a) with $+$0.085 (CI $+$0.073 to $+$0.097); the criterion therefore held (\cref{tab:suppl:mo}).
All three fitted read-outs in closed form by ridge regression on 512--1,024 labeled pool images and used 18--28 calls.
Against ME, the paired CIs on KonIQ lie below zero for all three ($-$0.092 to $-$0.015); on LIVE Challenge and BID, mo2's include zero.
We had predicted that one or two agents would keep the gap and gain less than ME on KonIQ; all three kept it.

\begin{table}[h]
  \centering
  \footnotesize
  \setlength{\tabcolsep}{2.5pt}
  \caption{\textbf{Agents under a sentiment constraint} (Qwen3.5-0.8B; SRCC, mean of 12 held-out prompts, real forward passes, so Base and ME$_1$ differ from the cached values of \cref{tab:main} by at most 0.002). Valence: ``good''$-$``poor'' log-probability gap between positive and negative sentences. QA: LLVisionQA accuracy change (points).}
  \label{tab:suppl:mo}
  \resizebox{\linewidth}{!}{%
  \begin{tabular}{llcccccc}
    \toprule
    & Edit & KonIQ & KADID & LIVE-C & BID & Valence & QA \\
    \midrule
    Base & -- & 0.643 & 0.548 & 0.693 & 0.662 & 6.5 & -- \\
    ME$_1$ & last layer, 1 neuron & 0.801 & 0.593 & 0.781 & 0.763 & 0.2 & $-$0.2 \\
    mo1 & layer 23, rank 1 & 0.747 & 0.590 & 0.758 & 0.730 & 7.0 & $-$0.5 \\
    mo2 & vision block 9, 1 neuron & 0.759 & 0.545 & 0.771 & 0.742 & 6.5 & $+$0.7 \\
    mo3 & layer 23, rank 1 & 0.728 & 0.590 & 0.744 & 0.721 & 6.5 & $-$0.4 \\
    \bottomrule
  \end{tabular}}
\end{table}

\paragraph{Cost.}
\Cref{tab:suppl:agentcost} lists, per run, the logged forward calls, wall-clock time and tokens.
Wall-clock time includes waiting for the shared GPU; the label-free runs took longer mainly because they ran concurrently and used almost their whole budget.

\begin{table}[t]
  \centering
  \small
  \setlength{\tabcolsep}{4pt}
  \caption{\textbf{Agent runs.} Calls: logged forward calls / budget. Tokens: output tokens and all input tokens processed (mostly cached context re-read on each step).}
  \label{tab:suppl:agentcost}
  \begin{tabular}{@{}lcrrr@{}}
    \toprule
    Run & Calls & Minutes & Output tok. & Input tok. \\
    \midrule
    run1 & 21 / 50 & 30 & 53k & 2.34M \\
    run2 & 14 / 50 & 28 & 57k & 2.38M \\
    run3 & 16 / 50 & 28 & 53k & 2.35M \\
    run4 & 13 / 25 & 24 & 40k & 1.83M \\
    run5 & 10 / 10 & 26 & 38k & 2.17M \\
    \midrule
    lf1--lf3 & 45--50 / 50 & 96--102 & 57--68k & 2.46--3.61M \\
    ll25, ll100 & 45, 46 / 50 & 25, 25 & 54k, 65k & 2.69M, 2.99M \\
    agiqa1--3 & 12--20 / 50 & 12--19 & 27--37k & 1.79--3.02M \\
    mo1--3 & 18--28 / 50 & 24--27 & 57--64k & 3.57--5.34M \\
    qa1--3 & 10--12 / 50 & 15--23 & 34--68k & 2.15--4.11M \\
    \bottomrule
  \end{tabular}
\end{table}
\section{Frozen Plans and Deviations}
\label{sec:suppl:plans}
Every experiment in this paper followed a plan written before any of its scores was seen: data and versions, the statistics, a decision rule where there was one, and our own predictions, kept apart from the rules.
In total we wrote 41 such plans.
Some early plans set thresholds for side effects; the paper reports side effects as measured, without these labels.
Predictions that failed are reported as such, \eg, we expected MTC refit on AGIQA-3K and R4-CGQA to stay within 0.02 of ME with no consistent winner, whereas ME was higher in 39 of 40 versions, significantly in 16 (\cref{sec:exp:main}).
Departures from a plan, all decided before the affected test scores were seen unless stated, were:
\begin{itemize}
  \item \emph{Offline check, neuron rescaling (0.8B).} The plan required offline scores to correlate with real forward passes at Pearson $\geq 0.99$; the unedited 0.8B model itself reached only 0.989 on one low-variance prompt, so the threshold for 0.8B became ``no lower than the unedited model'' (decided after the check, before any diagnostic score).
  \item \emph{Layer-probe analysis.} The first run selected a NaN layer (layer 0 has identical last tokens); after setting such layers to $-1$, only the analysis was rerun. Base, ME and last-layer probe numbers, which take no part in the selection, were unchanged.
  \item \emph{Undefined SRCC.} When an edited model gives the same one-word answer to every image, SRCC is undefined; the plan did not cover this, and we treat it as no gain (recomputing with SRCC $=0$ changes no conclusion).
  \item \emph{Engineering.} Batch sizes were reduced to fit memory (4B: 16, Gemma: 8); Gemma needed a wrapper for its image processor; 9B ran on a second machine and is not paired with the other models; where a later comparison mixed machines, the base model was rescored in the same run.
\end{itemize}
The 9B run on SPAQ and FLIVE (\cref{sec:suppl:newsets}) was planned ``if time allows'' and run last, with the same versions and criterion. Two limits apply throughout: the four test sets of \cref{tab:main} were consulted while developing the recipe, and several plans were written after related results were known; each plan lists such known biases.

%% file: sec/4_agents.tex
\section{Origin of the Structure: Agent Study}
\label{sec:agents}

\paragraph{Protocol.}
Five Claude agents (Claude Opus 5.5, model id \texttt{claude-opus-5-5}) worked in separate sessions on Qwen3.5-0.8B, each for about 30 minutes, on the same task: raise the SRCC of the good$-$poor score in \cref{eq:score} under the single prompt ``Rate the quality of the image.''
They could read and write the weights of any language or vision block and see the labeled 2,000-image KonIQ pool, but could run the model only through a logged tool, at most 256 images per call and 50 calls in total (25 and 10 for two of them).
Gradients and training loops were not allowed, and the agents were not told the thresholds we would use to judge them.
We scored each submitted edit once on the locked KonIQ test set, which no agent could see; each later audit followed a plan fixed before its scores were seen.

\begin{figure}[t]
  \centering
  \includegraphics[width=\linewidth]{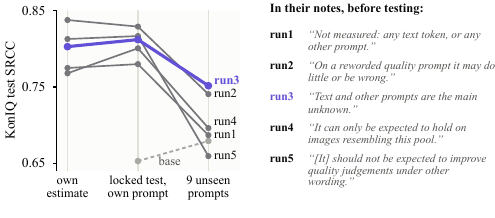}
  \caption{\textbf{Accurate on their own prompt, aware of its limit.}
    Left: for each agent editing Qwen3.5-0.8B, its own held-out estimate, the locked KonIQ test under its prompt, and the mean over nine prompts no agent saw (fit prompts written after the agents finished, disjoint from the 12 held-out prompts); run3, whose structure ME keeps, is highlighted.
    Right: lines from the agents' notes.}
  \label{fig:agents}
\end{figure}

\paragraph{Diagnosis first.}
Before editing, each agent fitted probes to hidden states on its own split of the pool and reached the conclusion of \cref{sec:gap} on its own: the score reached 0.59--0.63 SRCC, last-token probes 0.80--0.84, and, for the three that probed the vision encoder, 0.88--0.89.

\paragraph{The same remedy, for the same reason.}
All five then edited the last language layer, for the same stated reason: its output passes only through the final norm and the unembedding, so any edit can be evaluated exactly from cached states.
Each agent therefore compared 15--40 configurations offline and spent real forward passes mainly on capturing features and verifying its final edit.
The edits ranged from a single neuron (run2, run3) to rank-one updates of the down-projection (run1, run4, run5), but each wrote a label-fitted linear read-out into the good$-$poor direction.
They raised KonIQ test SRCC from 0.653 to 0.780--0.830, and four of the five predicted this within 0.011; the agent with 10 calls did as well as those with 50.
The convergence therefore reflects the protocol as much as the model: the agents share one model, one brief and one kind of budget, and are not independent samples.

\paragraph{They predicted their own limits.}
Every agent warned that its edit might work only for its own prompt, and our audit confirmed it: under nine prompts no agent saw (fit prompts 4--12; prompts 2 and 3 had been used in an earlier check the agents might have seen, and prompt 1 is theirs), the mean KonIQ gain of 0.16 shrank to between $-$0.02 and $+$0.07 (\cref{fig:agents}), and no edit met our transfer criterion (\cref{sec:exp:setup}).

\paragraph{From the agents to a recipe.}
Three further agents given the task without labels found no edit that raised SRCC by 0.03 on any dataset, so labels are needed to point the read-out at the information.
The agents contributed a diagnosis, a site and a structure; what they lacked was an objective that rewards transfer.
Nor does finding the site need an agent: with the same information and 50 calls, a scripted sweep that fits this single-neuron edit in every layer and keeps the best reaches 0.835 on the locked test, and the best of 45 random (layer, neuron) candidates 0.833, at least as high as the best agent and with similar gains under the nine unseen prompts (\cref{sec:suppl:agentruns}); layers 14--19 and 23--24 serve as well as the last in that search (agents' prompt, pool images; \cref{sec:gap} uses the 12 held-out prompts and test images).
ME keeps the single-neuron structure of their best-transferring edit, run3, and fits it to twelve prompts (\cref{sec:method}); \cref{fig:objective} compares the two objectives.

\paragraph{When the objective has a constraint.}
ME nearly erases the contrast between ``good'' and ``poor'' in text without images (\cref{sec:limits}): on 0.8B, the gap between positive and negative sentences falls from 6.5 to 0.2.
Three new agents received ME's information, the 12 fit prompts and the labeled pool, plus the requirement to keep this contrast and the model's other abilities, with the same budget and a text tool for sentences they wrote themselves; we scored each edit once under the 12 held-out prompts.
They did not use ME's site.
One turned an MLP neuron of vision block 9 into a quality read-out, which leaves text-only inputs unchanged by construction; two wrote rank-one read-outs into layer 23 whose effect on their own text-only sentences they penalized.
All three kept the contrast (6.5--7.0), factual answers (39 of 40, as the base model in the same evaluation, which answers 38 in \cref{tab:side}; greedy factual answers can differ in one of 40 questions between our two machines) and LLVisionQA accuracy within 0.7 points, and raised KonIQ SRCC from 0.64 to 0.73--0.76, significantly below ME's 0.80; two met all three of our criteria fixed before testing, which required a gain of at least 0.10 (\cref{tab:suppl:mo}).
The agents knew the contrast would be tested, whereas ME was not fitted for it.
Their direction, keeping text-only inputs out of the read-out, can be written into the closed form itself: projected onto the null space of text-only states, the edit (NME, \cref{sec:method}) keeps the contrast on the ten evaluation reviews (6.2--6.3) at ME's SRCC (0.80), above all three agents, though only 87\% of it after SST-2 sentences (\cref{sec:limits}).
So the agents supplied the diagnosis and the direction of a constraint, and the closed form absorbed both.

%% file: main.bib
@inproceedings{wu2024qbench,
  title     = {{Q-Bench}: A Benchmark for {General-Purpose} Foundation Models on Low-level Vision},
  author    = {Wu, Haoning and others},
  booktitle = {ICLR},
  year      = {2024}
}

@inproceedings{wu2024qinstruct,
  title     = {{Q-Instruct}: Improving Low-level Visual Abilities for Multi-modality Foundation Models},
  author    = {Wu, Haoning and others},
  booktitle = {CVPR},
  year      = {2024}
}

@inproceedings{wu2024qalign,
  title     = {{Q-Align}: Teaching {LMMs} for Visual Scoring via Discrete {Text-Defined} Levels},
  author    = {Wu, Haoning and others},
  booktitle = {ICML},
  year      = {2024}
}

@inproceedings{wu2024coinstruct,
  title     = {Towards Open-ended Visual Quality Comparison},
  author    = {Wu, Haoning and others},
  booktitle = {ECCV},
  year      = {2024}
}

@inproceedings{you2024depictqa,
  title     = {Depicting Beyond Scores: Advancing Image Quality Assessment through Multi-modal Language Models},
  author    = {You, Zhiyuan and others},
  booktitle = {ECCV},
  year      = {2024}
}

@inproceedings{zhu2024compare2score,
  title     = {Adaptive Image Quality Assessment via Teaching Large Multimodal Model to Compare},
  author    = {Zhu, Hanwei and others},
  booktitle = {NeurIPS},
  year      = {2024}
}

@inproceedings{you2025deqa,
  title     = {Teaching Large Language Models to Regress Accurate Image Quality Scores using Score Distribution},
  author    = {You, Zhiyuan and others},
  booktitle = {CVPR},
  year      = {2025}
}

@inproceedings{li2025qinsight,
  title     = {{Q-Insight}: Understanding Image Quality via Visual Reinforcement Learning},
  author    = {Li, Weiqi and others},
  booktitle = {NeurIPS},
  year      = {2025}
}

@inproceedings{wu2025visualqualityr1,
  title     = {{VisualQuality-R1}: {Reasoning-Induced} Image Quality Assessment via Reinforcement Learning to Rank},
  author    = {Wu, Tianhe and others},
  booktitle = {NeurIPS},
  year      = {2025}
}

@inproceedings{wang2023clipiqa,
  title     = {Exploring {CLIP} for Assessing the Look and Feel of Images},
  author    = {Wang, Jianyi and Chan, Kelvin C. K. and Loy, Chen Change},
  booktitle = {AAAI},
  year      = {2023}
}

@inproceedings{zhang2023liqe,
  title     = {Blind Image Quality Assessment via {Vision-Language} Correspondence: A Multitask Learning Perspective},
  author    = {Zhang, Weixia and others},
  booktitle = {CVPR},
  year      = {2023}
}

@inproceedings{chen2026qdebias,
  title     = {Mitigating Perception Bias: A {Training-Free} Approach to Enhance {LMM} for Image Quality Assessment},
  author    = {Chen, Baoliang and others},
  booktitle = {AAAI},
  year      = {2026}
}

@inproceedings{li2024aigiqa20k,
  title     = {{AIGIQA-20K}: A Large Database for {AI-Generated} Image Quality Assessment},
  author    = {Li, Chunyi and others},
  booktitle = {CVPRW},
  year      = {2024}
}

@inproceedings{zhang2025abench,
  title     = {{A-Bench}: Are {LMMs} Masters at Evaluating {AI-generated} Images?},
  author    = {Zhang, Zicheng and others},
  booktitle = {ICLR},
  year      = {2025}
}

@inproceedings{xu2023imagereward,
  title     = {{ImageReward}: Learning and Evaluating Human Preferences for {Text-to-Image} Generation},
  author    = {Xu, Jiazheng and others},
  booktitle = {NeurIPS},
  year      = {2023}
}

@inproceedings{kirstain2023pickapic,
  title     = {{Pick-a-Pic}: An Open Dataset of User Preferences for {Text-to-Image} Generation},
  author    = {Kirstain, Yuval and others},
  booktitle = {NeurIPS},
  year      = {2023}
}

@inproceedings{wang2024maagiqa,
  title     = {Large Multi-modality Model Assisted {AI-Generated} Image Quality Assessment},
  author    = {Wang, Puyi and others},
  booktitle = {ACM MM},
  year      = {2024}
}

@inproceedings{zhang2025qeval,
  title     = {{Q-Eval-100K}: Evaluating Visual Quality and Alignment Level for {Text-to-Vision} Content},
  author    = {Zhang, Zicheng and others},
  booktitle = {CVPR},
  year      = {2025}
}

@inproceedings{li2026r4cgqa,
  title     = {{R4-CGQA}: Retrieval-based Vision Language Models for Computer Graphics Image Quality Assessment},
  author    = {Li, Zhuangzi and others},
  booktitle = {CVPR},
  year      = {2026}
}

@inproceedings{hewitt2019control,
  title     = {Designing and Interpreting Probes with Control Tasks},
  author    = {Hewitt, John and Liang, Percy},
  booktitle = {EMNLP-IJCNLP},
  pages     = {2733--2743},
  year      = {2019}
}

@inproceedings{meng2022rome,
  title     = {Locating and Editing Factual Associations in {GPT}},
  author    = {Meng, Kevin and others},
  booktitle = {NeurIPS},
  year      = {2022}
}

@inproceedings{meng2023memit,
  title     = {{Mass-Editing} Memory in a Transformer},
  author    = {Meng, Kevin and others},
  booktitle = {ICLR},
  year      = {2023}
}

@inproceedings{mitchell2022mend,
  title     = {Fast Model Editing at Scale},
  author    = {Mitchell, Eric and others},
  booktitle = {ICLR},
  year      = {2022}
}

@inproceedings{dai2022neurons,
  title     = {Knowledge Neurons in Pretrained Transformers},
  author    = {Dai, Damai and others},
  booktitle = {ACL},
  year      = {2022}
}

@inproceedings{fang2025alphaedit,
  title     = {{AlphaEdit}: {Null-Space} Constrained Knowledge Editing for Language Models},
  author    = {Fang, Junfeng and others},
  booktitle = {ICLR},
  year      = {2025}
}

@inproceedings{gu2024harms,
  title     = {Model Editing Harms General Abilities of Large Language Models: Regularization to the Rescue},
  author    = {Gu, Jia-Chen and others},
  booktitle = {EMNLP},
  year      = {2024}
}

@inproceedings{cheng2023mmedit,
  title     = {Can We Edit Multimodal Large Language Models?},
  author    = {Cheng, Siyuan and others},
  booktitle = {EMNLP},
  year      = {2023}
}

@inproceedings{huang2024vlkeb,
  title     = {{VLKEB}: A Large {Vision-Language} Model Knowledge Editing Benchmark},
  author    = {Huang, Han and others},
  booktitle = {NeurIPS Datasets and Benchmarks},
  year      = {2024}
}

@inproceedings{ilharco2023taskarith,
  title     = {Editing Models with Task Arithmetic},
  author    = {Ilharco, Gabriel and others},
  booktitle = {ICLR},
  year      = {2023}
}

@inproceedings{li2023iti,
  title     = {{Inference-Time} Intervention: Eliciting Truthful Answers from a Language Model},
  author    = {Li, Kenneth and others},
  booktitle = {NeurIPS},
  year      = {2023}
}

@article{zou2023repe,
  title   = {Representation Engineering: A {Top-Down} Approach to {AI} Transparency},
  author  = {Zou, Andy and others},
  journal = {arXiv:2310.01405},
  year    = {2023}
}

@inproceedings{hu2022lora,
  title     = {{LoRA}: {Low-Rank} Adaptation of Large Language Models},
  author    = {Hu, Edward J. and others},
  booktitle = {ICLR},
  year      = {2022}
}

@article{alain2016probes,
  title   = {Understanding intermediate layers using linear classifier probes},
  author  = {Alain, Guillaume and Bengio, Yoshua},
  journal = {arXiv:1610.01644},
  year    = {2016}
}

@inproceedings{burns2023latent,
  title     = {Discovering Latent Knowledge in Language Models Without Supervision},
  author    = {Burns, Collin and others},
  booktitle = {ICLR},
  year      = {2023}
}

@inproceedings{orgad2025llmsknow,
  title     = {{LLMs} Know More Than They Show: On the Intrinsic Representation of {LLM} Hallucinations},
  author    = {Orgad, Hadas and others},
  booktitle = {ICLR},
  year      = {2025}
}

@article{fu2025hidden,
  title   = {Hidden in plain sight: {VLMs} overlook their visual representations},
  author  = {Fu, Stephanie and others},
  journal = {arXiv:2506.08008},
  year    = {2025}
}

@article{lu2024aiscientist,
  title   = {The {AI} Scientist: Towards Fully Automated {Open-Ended} Scientific Discovery},
  author  = {Lu, Chris and others},
  journal = {arXiv:2408.06292},
  year    = {2024}
}

@inproceedings{shaham2024maia,
  title     = {A Multimodal Automated Interpretability Agent},
  author    = {Rott Shaham, Tamar and others},
  booktitle = {ICML},
  year      = {2024}
}

@article{li2023agiqa,
  title   = {{AGIQA-3K}: An Open Database for {AI}-Generated Image Quality Assessment},
  author  = {Li, Chunyi and others},
  journal = {IEEE TCSVT},
  volume  = {34}, number = {8}, pages = {6833--6846},
  year    = {2024}
}

@article{zhang2023cgiqa,
  title   = {Subjective and Objective Quality Assessment for in-the-Wild Computer Graphics Images},
  author  = {Zhang, Zicheng and others},
  journal = {ACM TOMM},
  volume  = {20}, number = {4}, pages = {1--22},
  year    = {2023}
}

@article{hosu2020koniq,
  title   = {{KonIQ-10k}: An Ecologically Valid Database for Deep Learning of Blind Image Quality Assessment},
  author  = {Hosu, Vlad and others},
  journal = {IEEE TIP},
  volume  = {29}, pages = {4041--4056},
  year    = {2020}
}

@inproceedings{lin2019kadid,
  title     = {{KADID-10k}: A Large-scale Artificially Distorted {IQA} Database},
  author    = {Lin, Hanhe and Hosu, Vlad and Saupe, Dietmar},
  booktitle = {QoMEX},
  year      = {2019}
}

@article{ghadiyaram2016livec,
  title   = {Massive Online Crowdsourced Study of Subjective and Objective Picture Quality},
  author  = {Ghadiyaram, Deepti and Bovik, Alan C.},
  journal = {IEEE TIP},
  volume  = {25}, number = {1}, pages = {372--387},
  year    = {2016}
}

@article{ciancio2011bid,
  title   = {No-Reference Blur Assessment of Digital Pictures Based on Multifeature Classifiers},
  author  = {Ciancio, Alexandre and others},
  journal = {IEEE TIP},
  volume  = {20}, number = {1}, pages = {64--75},
  year    = {2011}
}

@inproceedings{fang2020spaq,
  title     = {Perceptual Quality Assessment of Smartphone Photography},
  author    = {Fang, Yuming and others},
  booktitle = {CVPR},
  year      = {2020}
}

@misc{qwen2026qwen35,
  title        = {{Qwen3.5}: Towards Native Multimodal Agents},
  author       = {{Qwen Team}},
  year         = {2026},
  howpublished = {\url{https://qwen.ai/blog?id=qwen3.5}},
  note         = {Official release blog post}
}

@article{gemmateam2026gemma4,
  title   = {{Gemma 4} Technical Report},
  author  = {{Gemma Team}},
  journal = {arXiv:2607.02770},
  year    = {2026}
}

@inproceedings{ying2020paq2piq,
  author = {Zhenqiang Ying and others},
  title = {From Patches to Pictures ({PaQ-2-PiQ}): Mapping the Perceptual Space of Picture Quality},
  booktitle = {CVPR},
  pages = {3575--3585},
  year = {2020}
}

@inproceedings{schuhmann2022laion,
  author = {Christoph Schuhmann and others},
  title = {{LAION-5B}: An Open Large-Scale Dataset for Training Next Generation Image-Text Models},
  booktitle = {NeurIPS Datasets and Benchmarks Track},
  year = {2022}
}

@misc{schuhmann2022aesthetic,
  author       = {Christoph Schuhmann},
  title        = {Improved Aesthetic Predictor},
  howpublished = {\url{https://github.com/christophschuhmann/improved-aesthetic-predictor}},
  year         = {2022}
}

@misc{qfuture2026qrealign,
  title        = {{Q-ReAlign} Models ({Mini-0.8B}, {Lite-4B}, {Pro-9B})},
  author       = {{Q-Future}},
  year         = {2026},
  howpublished = {\url{https://huggingface.co/q-future}},
  note         = {Hugging Face model collection}
}
